# Automated PD-L1 Tumor Proportion Scoring with Uncertainty-Guided Triage in Non-Small Cell Lung Cancer

Kesi Xu[1,*], Ayesha Azam[2], Tanvier Kaur[1], Monika Kondratowicz[3], Kinza Asim[2], Brinder Chohan[4], Paapa Appiah-Odame[5], Zhao Chen[6], David Snead[1,2], Nasir Rajpoot[1]

[1] Tissue Image Analytics Centre, Department of Computer Science, University of Warwick, Coventry CV4 7AL, UK

[2] Department of Pathology, University Hospitals Coventry and Warwickshire NHS Trust, Clifford Bridge Road, Coventry CV2 2DX, UK

[3] Department of Pathology, University Hospitals Birmingham NHS Foundation Trust, Mindelsohn Way, Edgbaston, Birmingham B15 2GW, UK

[4] Department of Histopathology, Royal Derby Hospital, Derby, UK

[5] University Hospitals Birmingham NHS Foundation Trust, Birmingham, UK

[6] School of Information and Intelligent Science, Donghua University, Shanghai, China

***Corresponding author:** Kesi Xu, Department of Computer Science, University of Warwick, Coventry CV4 7AL, UK. Email: kesi.xu@warwick.ac.uk

## Abstract

Programmed death-ligand 1 (PD-L1) tumor proportion score (TPS) informs immunotherapy selection in non-small cell lung cancer. Pathologists must distinguish tumor cells from stained immune cells and assess membranous staining in heterogeneous tissue. Automated scoring can provide a repeatable TPS estimate. Near the 1% and 50% cutoffs, the score should also indicate when closer review is warranted. We designed a three-stage framework to follow the sequence of TPS assessment. Stage 1 identifies tumor regions and excludes macrophage-rich patches. Stage 2, a dual-branch cell counting network (D2Count-Net), counts PD-L1-positive and PD-L1-negative tumor cells in these regions to calculate TPS. Stage 3, TPS-Guard, an uncertainty-guided triage mechanism, reports model uncertainty as the disagreement among independently trained models, refers the most uncertain slides, and flags scores near the 1% or 50% cutoff when model estimates disagree across that cutoff. Flags state the reason for referral or alert the pathologist to possible scoring errors on auto-reported slides. In a retrospective, single-center study, 154 whole-slide images were scored by three pathologists, and the model and each pathologist were compared in turn with the mean of the other two. On the 114 slides unseen in training, model mean absolute error (MAE) was 6.96 percentage points, within the range of the three pathologists (4.24 to 8.29), and model uncertainty correlated with absolute TPS error (Spearman $\rho = 0.77$) and inter-pathologist disagreement ($\rho = 0.71$). After referral of the 15% most uncertain slides, model and pathologist MAE both decreased. Applied without retraining to an external, publicly available dataset, TPS-Guard still directed review to higher-error cases, and referring more uncertain cases raised sensitivity at the 50% cutoff. TPS-Guard provides reasons for reviewing individual automated scores and identifies slides that also produce greater disagreement among pathologists. Prospective reader studies should assess TPS-Guard's effect on TPS reporting.

Programmed death-ligand 1 (PD-L1) expression in tumor cells informs the selection of pembrolizumab monotherapy in advanced non-small cell lung cancer (NSCLC).[1] PD-L1 expression is assessed by immunohistochemistry (IHC), with several antibody clones in use, including 22C3, 28-8, SP263 and SP142. 22C3, 28-8 and SP263 show broadly comparable tumor cell staining in NSCLC, whereas SP142 stains fewer tumor cells.[2] The development cohort in this study was stained with the VENTANA PD-L1 (SP263) assay. The external test set, from a public dataset, contained SP263-stained regions and regions stained with E1L3N, a clone used in laboratory-developed tests that have been evaluated against the 22C3 and SP263 assays.[3] Expression is reported as the tumor proportion score (TPS), the percentage of viable tumor cells with partial or complete membranous staining of any intensity.[4] The 1% and 50% thresholds inform treatment selection in the appropriate clinical context.[1,5] Scoring differences near these thresholds can therefore affect eligibility for a treatment option, particularly when the available material is a small biopsy.

Accurate TPS assessment depends on identifying viable tumor, separating tumor staining from staining in immune cells and macrophages, and recognizing any degree of membranous positivity including weak or incomplete staining.[4] Variable cellularity and heterogeneous staining further complicate estimation across a section. Agreement between pathologists varies across clinical cutoffs,[2,6–8] and differences between readers can exceed differences between assays.[6] Automated scores must therefore be evaluated alongside the variation already present in human assessment.

Automated TPS methods have used region-based or cell-based image analysis, including tumor-region segmentation and tumor-cell detection and classification.[9–14] Hondelink et al.[11] combined tumor segmentation with cell detection across whole-slide images (WSIs), while van Eekelen et al.[12] evaluated cell detection and TPS agreement across centers, assays, and scanners. Agreement across a cohort, however, does not identify which individual estimates need closer examination or explain why a score near a treatment threshold may be unreliable.

Studies have also examined how artificial intelligence (AI) could support pathologists in TPS assessment. In 584 tissue microarray samples, Ito et al.[15] found no statistically significant overall improvement with AI assistance, but reported higher scoring accuracy among cases with discordant pathologist scores. Makhlouf et al.[16] proposed referral for manual review when automated TPS fell within specified intervals near clinical cutoffs. Building on these approaches, we examine whether disagreement across models can help select TPS estimates for review, and whether cutoff-specific flags can give the reason for review.

Uncertainty estimation and selective classification offer a way to rank individual predictions for additional review.[17,18] For TPS, disagreement between models can signal instability in the numerical estimate. Proximity to a cutoff poses a separate problem: a small error may change the reported category even when models agree. A review strategy should therefore consider both variability in model predictions and evidence of possible threshold crossing.

We developed a framework that follows the sequence of TPS assessment: identifying the tumor area, estimating positive and negative tumor cell counts, and selecting cases for review. A dual-branch cell counting network combining detection and density estimation (D2Count-Net) accommodates variation in tumor cellularity. TPS-Guard uses disagreement across models to select slides for review and risk flags around the 1% and 50% cutoffs to record the reason for review. We evaluated associations with model error and disagreement among pathologists in 114 WSIs, and examined case selection in external regions of interest.

# 1. Materials and Methods

## 1.1 Study Cohort and Data Collection

### 1.1.1 Algorithm Development Cohort

This retrospective study at University Hospitals Coventry and Warwickshire (UHCW) used de-identified archival images under Research Ethics Committee (REC) approval, as detailed in Ethics Approval and Consent to Participate. The cohort comprised 154 PD-L1 IHC WSIs of NSCLC with paired hematoxylin and eosin (H&E) WSIs. The PD-L1 slides were stained with the VENTANA PD-L1 (SP263) Assay (Ventana Medical Systems, Inc., Tucson, AZ, USA) and digitized on a Philips Pathology Scanner SG300 (Philips Medical Systems Nederland B.V., Best, the Netherlands) at 40× magnification (0.25 μm/pixel). Table 1 summarizes the cohort characteristics.

Three anonymized pathology readers (P1, P2, P3) independently scored TPS for the 154 WSIs in the analysis cohort. The reader panel comprised pathologists with consultant-level National Health Service (NHS) diagnostic pathology experience in thoracic or surgical pathology and registrar-level surgical pathology experience, with varying levels of PD-L1 reporting experience.

Slides were reviewed using WASABI, a web-based slide viewer built on the Digital Slide Archive.[19] During scoring, pathologists could not see the other readers' scores, the algorithm output or their own previous TPS scores for the same case. They had access to the original diagnostic indication, but no other clinical information, treatment response data or outcomes. Following routine diagnostic practice, each pathologist reviewed the paired H&E and PD-L1 slides side by side and assigned a TPS from 0% to 100%. Range reports such as <1% were encoded as described in Supplementary Section S1. The algorithm used only the PD-L1 IHC slide.

The arithmetic mean of the three independent TPS scores (the three-reader mean) served as the reference for continuous scores and cutoff classifications. Three of P1's scores (WSIs 152, 371 and 400) were corrected by P1 after a check of slides with large differences between readers. All other scores are the original independent scores, and no consensus round was held. Results with the original scores are given in Supplementary Table S1. Section 1.11 defines the reader comparisons.

### 1.1.2 External Validation Cohort

The external cohort was used to test the model on a second staining clone and on other centers and scanners. External validation used the PD-L1 detection subset of the public IGNITE data toolkit described by Spronck et al.,[20] an expanded release of data previously used by van Eekelen et al.[12] The dataset comprised 344 regions of interest (ROIs) from 75 NSCLC patients at two centers: Radboud University Medical Center (RUMC), Nijmegen, the Netherlands, and Ospedale Sacro Cuore Don Calabria (SCDC), Negrar di Valpolicella, Italy. Slides were stained with E1L3N (Cell Signaling Technology, Inc., Danvers, MA, USA), 22C3 (Dako, Agilent Pathology Solutions, Carpinteria, CA, USA), or SP263 (Ventana Medical Systems, Inc., Tucson, AZ, USA). Images were acquired on three scanners. The main PD-L1 detection annotation file (pdl1_annotations.json) in version 1 of the IGNITE dataset contained 845,509 cell annotation records (Table 1). The ROI-only release limited evaluation to cell counting and excluded Stage 1. Unlike the prior WSI-level analysis, we derived a continuous TPS for each ROI from cell counts. We reserved 98 ROIs from 21 patients for testing (E1L3N, 53; SP263, 45); all three clones occurred in the training and validation ROIs, but the test ROIs contained no 22C3 staining.

| | UHCW (development) | RUMC/SCDC (external) |
|---|---|---|
| *Cohort overview* | | |
| Images, n | 154 WSIs | 344 ROIs (75 patients) |
| Training / validation partition, n | 42 cell-annotated WSIs (five outer folds, split by WSI) | 54 patients / 246 ROIs |
| Held-out test set, n | No separate test cohort; held-out fold-model estimates for 40 cell-annotated WSIs and 15-model ensemble estimates for 114 WSIs without cell-level annotations | 21 patients / 98 ROIs |
| Age, years (mean, range) | 70.3 (47–88) | Not reported‡ |
| Female sex, n (%) | 99 (65.1%) | Not reported‡ |
| *Specimen and staining* | | |
| Specimen type | Surgical resection 69 (44.8%); lung wedge biopsy 59 (38.3%); endobronchial ultrasound (EBUS)-guided biopsy 25 (16.2%); lymph node 1 (0.6%) | Biopsy, resection, and tissue microarray cores; per-type counts in IGNITE Figure 2B‡ |
| PD-L1 antibody clone | SP263 (Ventana) | RUMC: E1L3N (Cell Signaling Technology) and 22C3 (Agilent); SCDC: 22C3 (Agilent) and SP263 (Ventana) |
| Scanner and resolution | Philips Pathology Scanner SG300; scanned at 40× (0.25 $\mu m \cdot px^{-1}$), analyzed at 20× (0.5 $\mu m \cdot px^{-1}$) | 3DHistech Pannoramic 1000 and Pannoramic 250 Flash II (RUMC), scanned at 40× (0.25 $\mu m \cdot px^{-1}$); Roche Ventana iScan HT (SCDC), scanned at 20× (0.5 $\mu m \cdot px^{-1}$); all analyzed at 20× |
| Evaluation level | Whole-slide image | Region of interest |
| *TPS distribution, n (%)* | | |
| <1% | 70 (45.5%) | 44 (44.9%) |
| 1% to <10% | 35 (22.7%) | 9 (9.2%) |
| 10% to <50% | 28 (18.2%) | 6 (6.1%) |
| ≥50% | 21 (13.6%) | 39 (39.8%) |
| *Annotations* | | |
| Analysis cohort | 154 PD-L1 WSIs | 98 ROIs (held-out test) |
| Expert TPS scores | 154 WSIs (3 pathologists) | Derived from cell counts |
| Tumor region polygons | 30 WSIs | Not applicable |
| Cell-level annotations | 42 WSIs, 40 of them in the analysis cohort (50,615 nuclei across PD-L1-positive tumor, PD-L1-negative tumor, and non-tumor classes) | 344 ROIs (845,509 annotation records across PD-L1+ tumor, PD-L1− tumor, and 'non-tumor' classes) |

**Table 1.** UHCW development and RUMC/SCDC external validation cohorts. In UHCW, age and sex describe 152 patients, two of whom contributed two WSIs each. Specimen type and TPS categories describe 154 WSIs. In RUMC/SCDC, TPS categories describe the 98 held-out test ROIs. ‡Demographics not reported in the source dataset; RUMC/SCDC cohort statistics were calculated from the metadata and annotation files in version 1 of the IGNITE dataset described by Spronck et al.[20].

## 1.2 Overview of the Proposed Framework

The framework has three stages (Figure 1). Stage 1 identifies tumor regions and excludes patches classified as non-tumor or macrophage-rich. Stage 2 applies D2Count-Net to 224 × 224-pixel tumor patches at 20× magnification. A scene classifier estimates probabilities of background (0 tumor cells), low (1–14), medium (15–39), and high (≥40) cellularity, which determine the weights of detection and density estimates before slide-level aggregation. Stage 3, TPS-Guard, refers the 15% of slides with the highest model uncertainty for pathologist review and applies three flags. Each flag records its reason. Slides that are not referred are auto-reported.

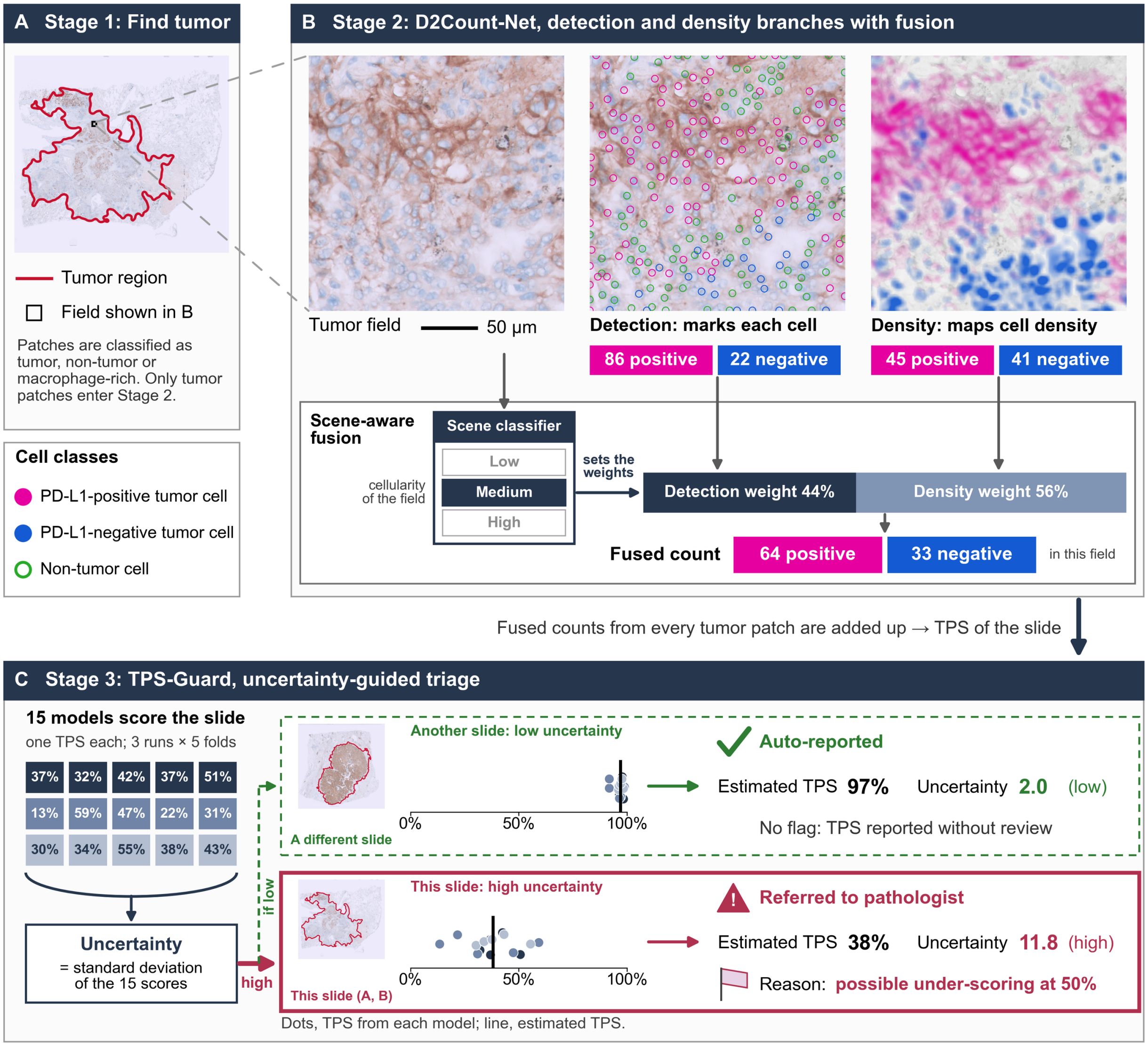

**Figure 1.** Overview of the three-stage TPS estimation framework, shown on one slide. (A) Stage 1, tumor region classification with a linear classifier on Virchow2 features (tumor, non-tumor, and macrophage-rich tissue; macrophage-rich regions are merged into non-tumor); red outline, predicted tumor region; box, field shown in (B). (B) Stage 2, D2Count-Net cell analysis of one tumor field. The detection branch marks individual cells and the density branch maps cell density; each gives a count of PD-L1-positive and PD-L1-negative tumor cells. A scene classifier assigns a cellularity class, which sets the weights of the two counts in the fused count; the field spans four patches, so the weights shown are their mean. Images show the mean output of the 15 models, rendered with overlapping windows. Fused counts from all tumor patches are added up to give the slide-level TPS. (C) Stage 3, TPS-Guard uncertainty-guided triage. Fifteen independently trained models each give a TPS; the estimated TPS is calculated from their mean cell counts, and the uncertainty is the standard deviation (SD) of the 15 TPS values. The 15% most uncertain slides are referred to a pathologist and the others are reported automatically; a flag gives the reason for referral or an advisory label on an auto-reported slide. The slide in (A) and (B) is referred; a second slide with low uncertainty is shown for comparison. Dots, TPS from each model; line, estimated TPS. Scale bar, 50 µm.

### 1.3 Annotation and Algorithm Training Data

Expert annotations comprised two categories: tumor region polygons on 30 WSIs (Section 1.5) and cell-level nucleus annotations on 42 WSIs (50,615 nuclei: PD-L1-positive tumor, PD-L1-negative tumor, and non-tumor cells; Section 1.6). From the annotated regions, 1,115 non-overlapping patches (224 × 224 pixels, 20×) were extracted. Of these, 235 contained no tumor cells, reflecting tumor–stroma boundary patches. The 42 WSIs were split into five outer folds by slide, and each patch was evaluated by the fold models that had not seen its source slide.

Forty of the cell-annotated WSIs are also in the analysis cohort of 154 WSIs. Each of these 40 WSIs was scored by the three fold models, one per training run (Section 1.4), that excluded it from training and validation. The other 114 WSIs were not used for cell-level training. We report two evaluation frames: the full cohort of 154 WSIs and the 114 WSIs without cell-level annotations, which form the primary analysis. These model assignments determine the final TPS estimate; model uncertainty and patch-derived risk features use outputs from all 15 cross-validation models, as specified in Supplementary Section S2. For each of the 40 cell-annotated WSIs, 12 of these 15 models used its annotated regions for training or validation, so uncertainty-based referral is fully held out only in the 114 WSIs.

### 1.4 External Validation and Experimental Design

The cell analysis model was evaluated on the external test set under three conditions: direct application without retraining (zero-shot transfer), fine-tuning on the external cohort, and external-only training. Three independent training runs on the development cohort each produced five cross-validation models, with WSIs kept separate between training and validation within each fold. Each run supplied models for both direct application and fine-tuning. External-only training used one ensemble of five models. Results for direct application and fine-tuning are reported for the 15 models, with cell counts averaged across models. The external training, validation, and test partitions contained 185 ROIs from 44 patients, 61 ROIs from 10 patients, and 98 ROIs from 21 patients, respectively, with no patient overlap.

1. Application without retraining (zero-shot transfer). The development-trained models were applied directly to the external test ROIs without adaptation.
2. Fine-tuning. Each development-trained model was adapted to the external training/validation split using the same multi-task loss. Cell counts from the 15 adapted models were averaged. The 15 models shared one external training/validation partition and differed in development-cohort initialization.
3. External-only training. Five models were trained independently on the external training/validation split, starting from ImageNet-pretrained encoder weights and using the same architecture as the development-cohort models. Cell counts were averaged across the five models and evaluated on the same held-out ROI test set.

For each external ROI, TPS was calculated from counts aggregated across non-overlapping patches of 224 × 224 pixels. Training settings and TPS calculation details are given in Supplementary Sections S3.2 and S2.

### 1.5 Stage 1: Tumor Region Segmentation

Virchow2[21], a vision transformer pretrained on histopathology images, was frozen as the patch-level feature extractor. On 30 WSIs, annotations marked invasive tumor, non-tumor (stroma, normal lung, and other benign elements), and macrophage-rich intratumoral regions. The third class captures PD-L1-positive macrophages that can mimic positive tumor cells. A linear classifier on the Virchow2 class token features assigned each 224 × 224-pixel patch at 20× to one of these three classes. At inference, patches classified as macrophage-rich were merged into non-tumor and excluded from Stage 2. Because exclusion operates at patch level, tumor cells within mixed patches may also be omitted.

### 1.6 Stage 2: D2Count-Net Cell Analysis

Within the segmented tumor region, D2Count-Net processes non-overlapping 224 × 224-pixel patches at 20× magnification. A shared EfficientNet-B0 encoder[22] feeds the detection and density branches, each implemented as a U-Net-style decoder.[23] The branches provide two estimates of cell counts for tissue fields with different cellularity. Pathologists annotated 50,615 nuclei on 1,115 patches from 42 WSIs, of which 3,152 were PD-L1-positive tumor cells (6.2%), 18,060 PD-L1-negative tumor cells and 29,403 non-tumor cells (Supplementary Figure S4).

The detection branch marks individual cells and classifies each as PD-L1-positive tumor, PD-L1-negative tumor, or non-tumor. This branch is designed to count cells that can be resolved individually. The density branch predicts continuous density maps for positive and negative tumor cells, with counts obtained by spatial integration. This branch provides a complementary estimate for crowded fields where individual cells are difficult to separate.

A tissue scene classifier estimates the probability that each patch belongs to one of four tumor cellularity categories: background, low, medium, or high. A fixed mapping converts these probabilities into weights for the detection and density branches. This weighting favors detection in sparsely cellular fields and density estimation in crowded fields. Supplementary Section S3.1 gives the equations and fixed priors, and Supplementary Figure S2 illustrates the architecture and fusion.

**1.7 Slide-Level TPS Calculation**

Slide-level TPS is computed by aggregating cell counts across all tumor patches:

$$TPS_{slide} = \frac{\Sigma\, N_{p,i}}{\Sigma\,(N_{p,i} + N_{n,i})} \times 100\% \qquad (1)$$

where the summation is over all patches $i$ within the segmented tumor region, and $N_{p,i}$ and $N_{n,i}$ are the fused positive and negative tumor cell counts, respectively. Each of the 40 cell-annotated WSIs is scored by its three held-out fold models, one from each training run. For the other 114 WSIs, positive and negative tumor cell counts are averaged across the 15 models before TPS is calculated from these averages. Non-tumor cells (immune cells, stroma, and macrophages) are excluded from both numerator and denominator, consistent with the PD-L1 TPS definition.

**1.8 Uncertainty Quantification and Referral**

Model uncertainty was measured as the standard deviation (SD) of TPS estimates from the 15 independently trained cross-validation models. This SD describes disagreement among model estimates for a given slide. We retrospectively assessed the association of model uncertainty with absolute TPS error and inter-reader disagreement. Reader scores are not required to calculate model uncertainty. Supplementary Section S2 details the TPS calculations used for development WSIs and external ROIs.

Cases are ranked by model uncertainty, with the highest values referred for pathologist review. The referral rate is the proportion of slides referred for pathologist review. In the proposed workflow, the pathologist remains responsible for the final TPS report for every slide.

### 1.9 Fusion Weights

The four scene priors and their mapping to the detection weight were fixed across the development and external cohorts. We assessed the contribution of scene-based weighting by recalculating TPS from the same branch outputs with equal weights for the detection and density branches (Supplementary Section S3).

### 1.10 Stage 3: Uncertainty-Guided Triage with TPS-Guard

TPS-Guard adds cutoff-specific flags to the uncertainty ranking described in Section 1.8. A small numerical error near the 1% or 50% threshold can change the reported TPS category, even when model uncertainty is low.

TPS-Guard gives three reasons to review an automated score. The flag for possible over-scoring at 1% applies to fused TPS from 1% to less than 5% when model estimates straddle 1%, the fused estimate falls within a fixed margin above 1%, or predicted positive tumor cells are few or concentrated in a few patches; such a slide may be negative at the 1% cutoff. The flag for possible under-scoring at 50% applies to fused TPS from 20% to less than 50% when a component or alternative estimate reaches 50%, or high-TPS patches account for a substantial share of predicted tumor cells; such a slide may have a TPS of 50% or more. The flag for limited tumor-cell sampling identifies scores based on few predicted tumor cells or scoring patches, with additional rules for specimen type and proximity to a clinical cutoff. Supplementary Section S2 specifies the full criteria.

TPS-Guard selects the 15% of slides with the highest model uncertainty for review. On these referred slides, the flags give the reason for review; on the auto-reported slides, they are reported as advisory labels, and a flag never forces referral. Exact ranking and rounding rules are given in Supplementary Section S2. The three reasons (possible over-scoring at 1%, possible under-scoring at 50%, or limited tumor-cell sampling) can be considered alongside the TPS estimate and the slide. The rules were derived on the development cohort using model outputs and specimen type.

Sensitivity analyses on the 114 WSIs without cell-level annotations and confidence intervals are reported in Supplementary Section S6 and Supplementary Table S2. Supplementary Section S7 reports matched-slide reader comparisons and a score-substitution simulation for uncertainty ranking (the main analysis) without flags.

### 1.11 Statistical Analysis

Agreement between the model and the reference was summarized by mean absolute error (MAE) in TPS percentage points, the intraclass correlation coefficient for absolute agreement (two-way random effects, single measures; ICC(2,1)),[24] and Bland–Altman analysis.[25] At the 1% and 50% cutoffs, sensitivity and specificity were the proportions of reference-positive and reference-negative cases correctly classified, respectively. Three-tier accuracy was the proportion assigned to the correct TPS category (<1%, 1% to <50%, or ≥50%).

On the development cohort, estimated TPS was compared with the mean of three readers. Two reader benchmarks were used. The pairwise benchmark averaged agreement across the three pairs of pathologists. In the leave-one-reader-out benchmark, the model and the held-out pathologist were each compared with the mean of the remaining two pathologists, so that both were judged against the same reference. Reader comparisons were recalculated for every set of slides analyzed.

The associations of model uncertainty with absolute model error and with inter-pathologist disagreement (the population SD of the three reader scores) were assessed by Spearman correlation, with two-sided P values. One-sided permutation tests (10,000 permutations) assessed whether flagged slides had higher rates of the specified error events than all 114 WSIs without cell-level annotations. Each flag was evaluated at its own cutoff: for possible over-scoring at 1%, a reference TPS below 1% with an estimated TPS of 1% or more; for possible under-scoring at 50%, a reference TPS of 50% or more with an estimated TPS below 50%. P values were reported without adjustment for the two permutation tests.

Ninety-five percent confidence intervals (CIs) were obtained by percentile bootstrap resampling (2,000 resamples unless stated otherwise). Resampling used slides in the development cohort and patients in the external cohort, retaining all ROIs from each sampled patient. The external comparison of fine-tuning with application without retraining used a paired patient bootstrap (10,000 resamples). Training and resampling settings are given in Supplementary Section S3.2.

# 2. Results

## 2.1 Model–Expert Agreement

Slides auto-reported after referral of the 15% most uncertain slides had lower TPS error and less reader disagreement than all 114 slides. Before referral, model MAE against the three-reader mean was 6.70 percentage points (95% CI 4.66 to 8.97) across the 114 WSIs without cell-level annotations; the mean pairwise difference between pathologists was 6.41. Against the mean of the other two readers, the three pathologists had MAEs of 4.24, 5.49 and 8.29 points and the model 6.96, a difference from a single pathologist of 0.95 points (95% CI −0.53 to 2.48), about one-seventh of the mean disagreement between pathologists (6.41 points). At the 50% cutoff the model and a single pathologist placed 5.6% and 6.1% of slides in a different category from the mean of the other two readers; at the 1% cutoff, 19.9% and 17.5%. TPS-Guard selected 17 slides (14.9% of 114 WSIs) with the highest model uncertainty for review. On the 97 auto-reported slides, model MAE against the three-reader mean was 4.96 points and the mean pairwise difference between pathologists 4.93; against the mean of the other two readers, the model (5.18) and a single pathologist (4.61) differed by 0.57 points (95% CI −0.71 to 2.03; Table 2; Supplementary Table S1). On these slides the model placed fewer slides than a single pathologist in a different category from the reference at the 50% cutoff (1.7% vs 3.1%) and more at the 1% cutoff (23.0% vs 19.6%). In the full development cohort of 154 WSIs, referral also reduced model MAE, but on the auto-reported slides model MAE was higher than the pairwise pathologist MAE (Table 2). On all 154 WSIs, which include the 40 used for cell-level training, the model's error against the mean of the other two readers was 6.79 points and a single pathologist's 5.67 (difference 1.12, 95% CI −0.06 to 2.42).

Before referral, predictions followed reader scores across the TPS range but tended to underestimate high TPS (Figure 2; Supplementary Figure S1); the ten largest residuals are listed in Supplementary Table S5. Mean bias was −5.0 percentage points (95% limits of agreement −29.2 to 19.2; Supplementary Figure S1I).

*Panel A. Baseline vs TPS-Guard cutoff performance after referral of the 15% most uncertain slides.*

| Configuration | Referred | Pathologist MAE (pp) | Model MAE (pp) | Model ICC | Sensitivity at 50% (n/N) | Specificity at 1% | Three-tier accuracy |
|---|---|---|---|---|---|---|---|
| *WSIs unseen in training (n = 114)* | | | | | | | |
| Baseline, no referral | 0.000 (0/114) | 6.41 | 6.70 | 0.862 | 0.63 (10/16) | 0.755 | 0.737 |
| 15% most uncertain referred | 0.149 (17/114) | 4.93 | 4.96 | 0.762 | 0.50 (2/4) | 0.755 | 0.732 |
| *Full development cohort (n = 154)* | | | | | | | |
| Baseline, no referral | 0.000 (0/154) | 6.13 | 6.56 | 0.869 | 0.67 (14/21) | 0.757 | 0.740 |
| 15% most uncertain referred | 0.149 (23/154) | 4.57 | 5.07 | 0.767 | 0.60 (3/5) | 0.757 | 0.733 |

*Panel B. Leave-one-reader-out benchmarks on the cohort of 114 WSIs without cell-level annotations.*

| Held-out reader | Pathologist MAE (pp) | Model MAE (pp) | Pathologist ICC | Model ICC |
|---|---|---|---|---|
| Average across hold-outs | 6.01 | 6.96 | 0.896 | 0.853 |
| P1 held out | 5.49 | 6.17 | 0.912 | 0.886 |
| P2 held out | 4.24 | 6.42 | 0.952 | 0.876 |
| P3 held out | 8.29 | 8.28 | 0.825 | 0.797 |

**Table 2.** Model and pathologist agreement with and without TPS-Guard. (A) WSIs unseen in training: the 114 WSIs without cell-level annotations, which no model used for training or validation. Full development cohort: all 154 WSIs, including the 40 cell-annotated WSIs, whose TPS comes from the fold models that excluded them but whose uncertainty uses all 15 models, 12 of which used their annotated regions for training or validation. Referred is the fraction of slides referred for pathologist review. After referral, metrics are computed on the auto-reported slides only; ICC is lower on these slides because their TPS range is narrower, and pairwise pathologist ICC decreases on the same slides (Supplementary Table S4). Model metrics use the three-reader mean; pathologist MAE averages the three pairwise differences on each auto-reported set. Sensitivity at 50% (n/N) gives correctly classified slides with reference TPS ≥50% divided by all auto-reported slides with reference TPS ≥50%. Denominators vary with the auto-reported set.

(B) Common-reference analysis (114 WSIs without cell-level annotations). The model and each held-out pathologist are compared with the mean of the other two readers.

WSI, whole-slide image; TPS, tumor proportion score; MAE, mean absolute error in percentage points (pp); ICC, intraclass correlation coefficient (2,1); Three-tier accuracy, accuracy for <1%, 1% to <50%, and ≥50% TPS.

In the common-reference analysis of the 114 WSIs without cell-level annotations, the model and each held-out reader were compared with the mean of the other two readers. Average model and reader ICCs were 0.853 and 0.896, respectively (Table 2, Panel B). Each pathologist had a higher

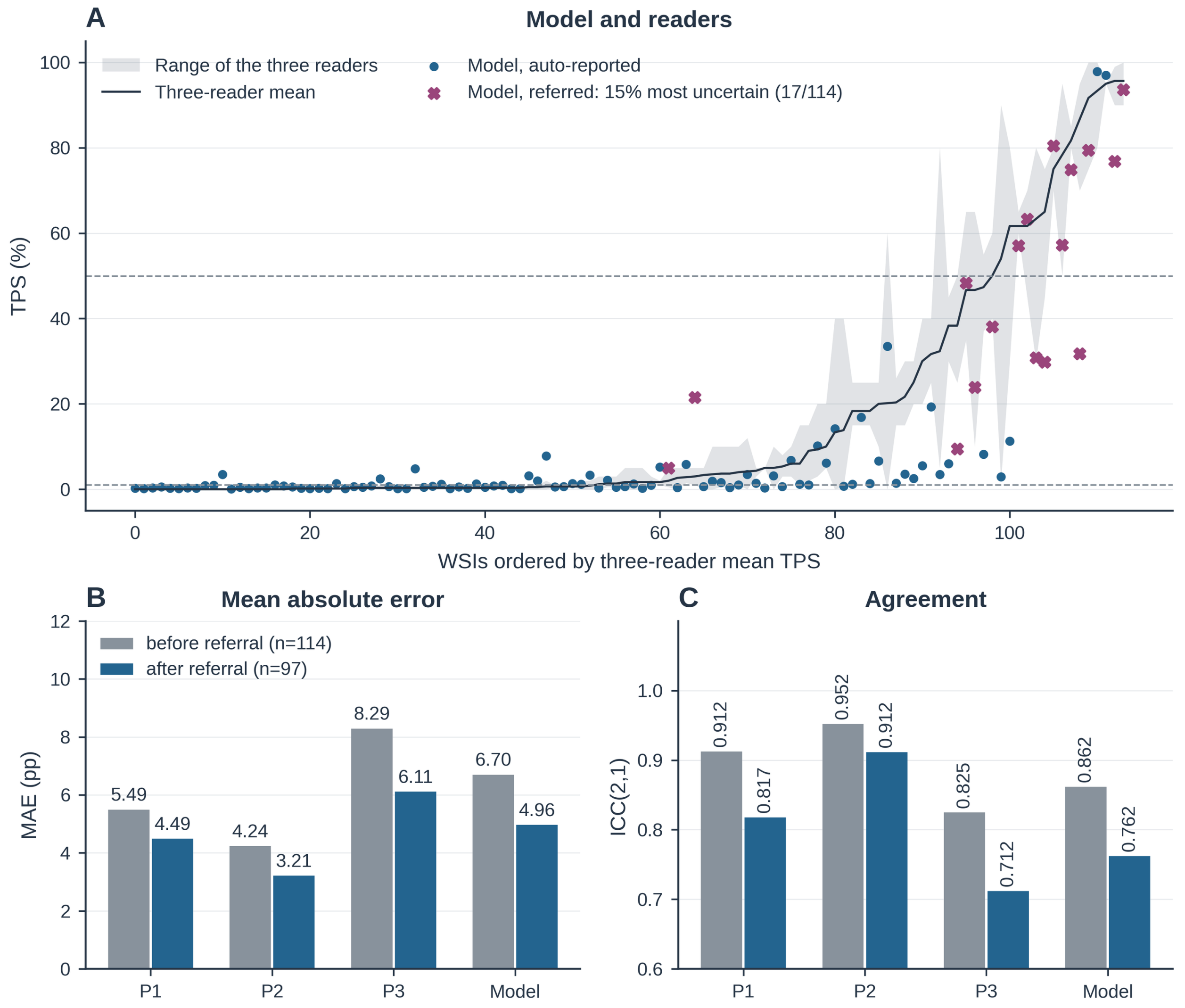

**Figure 2.** Model–expert agreement on the 114 WSIs without cell-level annotations before and after referral of the 15% most uncertain slides (97 of 114 auto-reported; Table 2, Panel A, row 2). (A) All 114 WSIs ordered by three-reader mean TPS; gray band, range of the three reader scores; blue points, estimated TPS on auto-reported slides; magenta crosses, referred slides; dashed lines, 1% and 50% cutoffs. (B) MAE before and after referral for each pathologist, compared with the mean of the other two, and for D2Count-Net, compared with the three-reader mean. (C) ICC for the same comparisons. Cohort-level operating characteristics across referral rates are reported in Figure 5; numerical values are summarized in Table 2. MAE, mean absolute error; ICC, intraclass correlation coefficient.

ICC than the model, and P1 and P2 had lower MAE; P3 and the model had MAEs of 8.29 and 8.28 percentage points. Reader agreement was lowest in the 19 endobronchial ultrasound (EBUS)-guided biopsies, where the model's error against the other two readers was 1.36 points higher than a single pathologist's (95% CI −2.88 to 6.81; Supplementary Table S6).

### 2.2 Tumor Segmentation and Cell-Counting Outputs

The segmentation overlays showed overlap between predicted tumor regions and pathologist annotations in surgical resection and EBUS-guided biopsy specimens (Supplementary Figure S3).

Figure 3 shows the output of each branch beside the pathologist annotation in three annotated fields of low, medium, and high cellularity. Where tumor cells were well separated, the detection branch counted most of the annotated tumor cells. In crowded fields it counted fewer tumor cells than the pathologist, whereas the density count stayed close to the annotated count and the fused count lay between the two.

In the external ROI example, fine-tuning brought the detection estimate of PD-L1-negative cells closer to the annotated count, while density and fusion overestimated that count (Supplementary Figure S7).

### 2.3 Branch Complementarity Analysis

Fusion had lower overall TPS error than either detection or density alone, whose MAEs were 7.09 and 8.27 percentage points, respectively (Figure 4). The most accurate branch varied with TPS. In per-slide comparisons, density most often had the lowest error below 10%, and detection at 10% or above (Figure 4B).

Fusion gave the highest accuracy at the 1% threshold (78.9%); at 50% TPS, its accuracy was 94.7% (Figure 4C). The corresponding accuracies were 73.7% and 95.6% for detection, and 76.3% and 92.1% for density. Detection gave 80.3% sensitivity and 66.0% specificity at the 1% cutoff. Density preserved specificity at this cutoff but missed more reference-positive slides than fusion.

### 2.4 Model Uncertainty, Reader Disagreement, and Case Selection

Model uncertainty correlated with absolute TPS error (Spearman $\rho = 0.77$; 95% CI 0.66 to 0.83; $P < 10^{-22}$; Figure 5C). Model uncertainty correlated with the spread of the three reader scores, Spearman $\rho = 0.71$ (95% CI 0.60 to 0.79; Figure 5D). Slides on which the model was uncertain therefore tended to be the slides on which the pathologists disagreed. Both intervals were estimated by WSI-level bootstrapping. The two slides with the highest model uncertainty in the cohort were two of the three slides whose score was corrected. Mean absolute TPS error increased from 0.6 percentage points in the lowest uncertainty quartile to 16.4 in the highest (Supplementary Figure S6B).

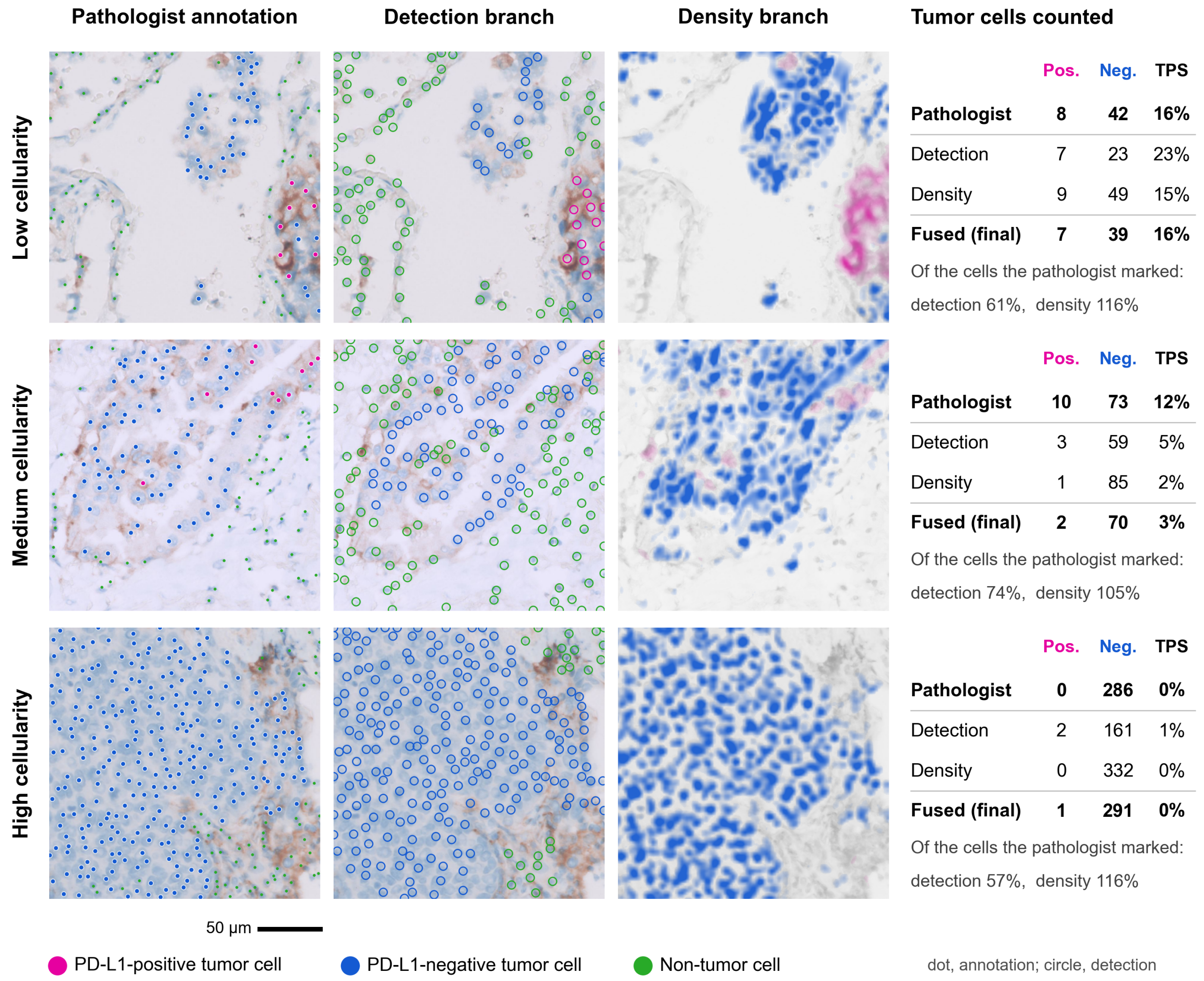

**Figure 3.** Cell-counting outputs in annotated fields. Three fields (224 × 224 µm) of low, medium, and high tumor cellularity from WSIs with cell-level annotations. Left, pathologist annotation (dots). Middle, detection branch; circles mark the cells called by one model. Right, density branch; shading shows the estimated density of PD-L1-positive (magenta) and PD-L1-negative (blue) tumor cells. The table gives the tumor-cell counts and the TPS of the field from the annotation, from each branch, and from the fused output. Model counts are means of the three models that did not use the slide in training. The fields are examples chosen to show the range of cellularity. The detection and density images were computed with overlapping windows for display; the counts come from the standard non-overlapping patches. Scale bar, 50 µm.

Uncertainty ranking alone left an auto-reported set with lower model error and less reader disagreement. With 10% of slides referred, model MAE was 5.48 percentage points. Mean pairwise pathologist MAE was 5.28 percentage points on the auto-reported slides, compared with

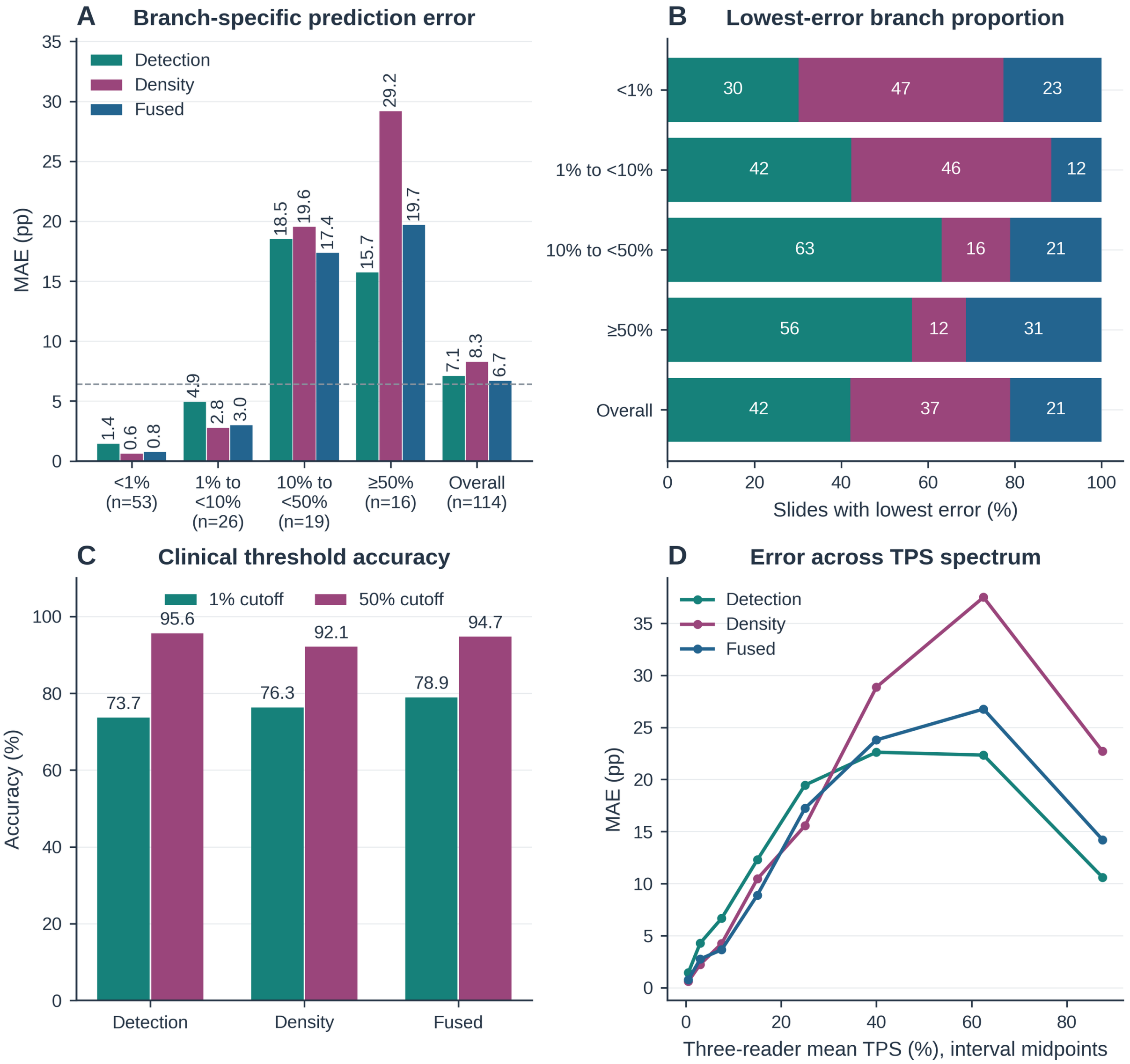

**Figure 4.** Branch complementarity on the 114 WSIs without cell-level annotations. (A) TPS MAE by reference TPS range for the detection branch, the density branch and the fused output; the dashed line marks the overall mean pairwise inter-pathologist MAE (6.4 percentage points) for reference. (B) Per-case branch with the lowest absolute TPS error. (C) Accuracy at the 1% and 50% clinical thresholds. (D) Mean absolute TPS error across intervals of reference TPS. Points show the mean error within each interval, plotted at its midpoint and connected to aid visualization. Interval boundaries are 0%, 1%, 5%, 10%, 20%, 30%, 50%, 75% and 100%; each displayed interval contains at least three slides. MAE, mean absolute error.

6.41 across all 114 slides (Supplementary Section S7). In the leave-one-reader-out benchmark, model MAE on auto-reported slides exceeded that of a single pathologist by 0.07 to 1.17 percentage points with up to 50% of slides referred and was lower with 55% or more referred; both decreased as more uncertain slides were referred (Figure 5A). At the 1% cutoff, the model placed more auto-reported slides in a different category from the reference than a single pathologist at every referral rate; at the 50% cutoff, it placed fewer with up to 25% of slides referred, and neither placed any with 30% or more referred (Figure 5B).

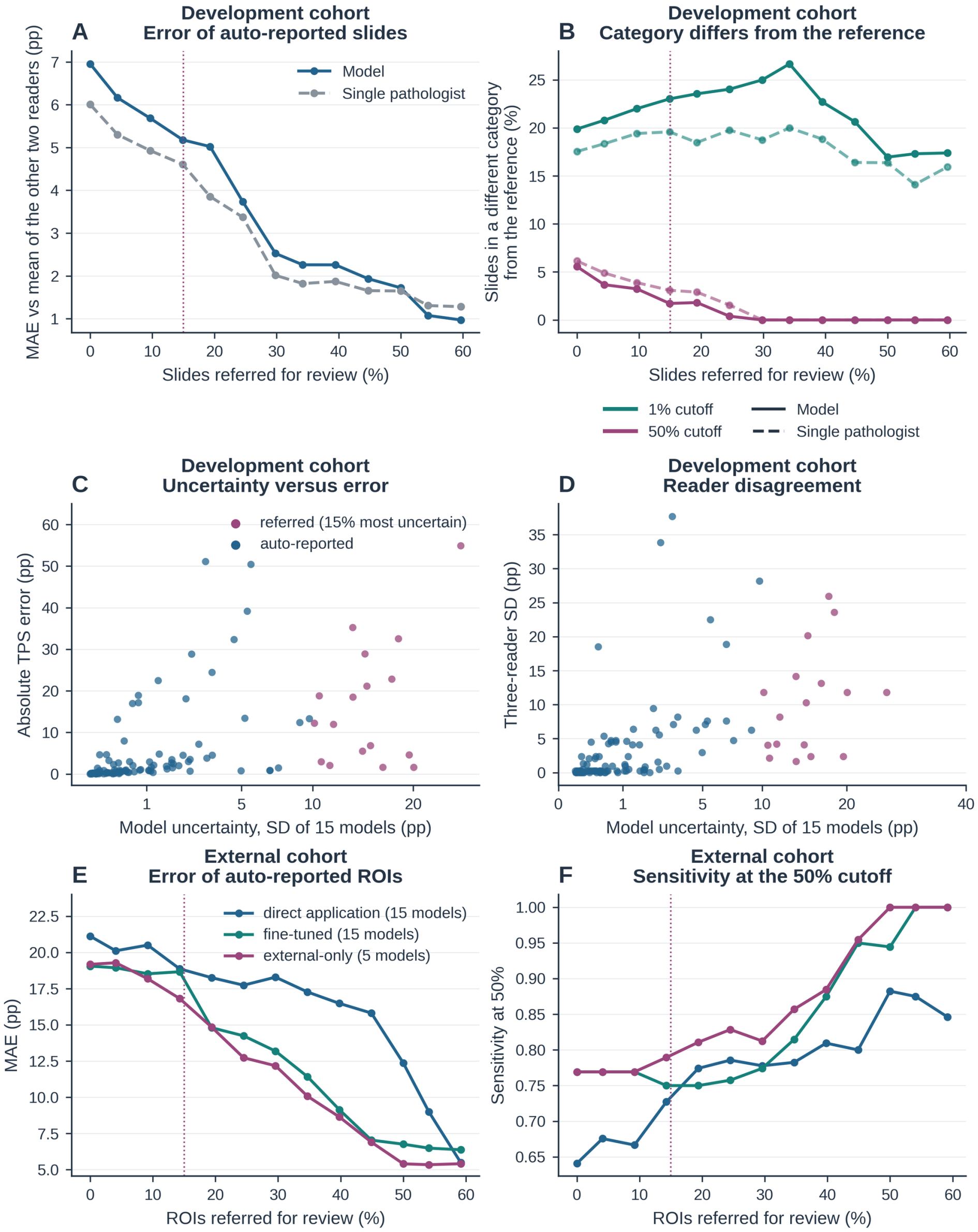

**Figure 5.** Model uncertainty, reader disagreement and referral rate. Uncertainty is the SD of TPS across the 15 models (five for the external-only ensemble). (A, B) 114 WSIs without cell-level annotations, leave-one-reader-out: the model and each held-out pathologist are compared with the mean of the other two readers, and values are averaged over the three held-out pathologists. (A) MAE of auto-reported slides against the percentage of slides referred; solid line, model; dashed line, single pathologist (the held-out pathologist). (B) Percentage of auto-reported slides placed in a different category from the reference at the 1% cutoff (teal) or the 50% cutoff (magenta); line styles as in A. (C, D) Uncertainty versus absolute TPS error and versus the SD of the three reader scores; magenta, the 15% most uncertain slides (referred); horizontal axes use square-root scales. (E, F) External cohort (98 ROIs): MAE and sensitivity at 50% TPS for the 15-model ensemble applied without retraining, the fine-tuned 15-model ensemble and the external-only 5-model ensemble. Slides or ROIs with the highest uncertainty are referred first; the dotted line marks 15% referred. MAE, mean absolute error; SD, standard deviation; pp, percentage points.

The cutoff-specific flags identified groups with higher rates of their target errors than all 114 WSIs without cell-level annotations. In exploratory analyses of the development cohort used to define the rules, the flag for possible over-scoring at 1% identified false positives at 1% in 7 of 21 flagged slides (33.3%; 95% CI 14.3% to 55.2%), compared with a cohort rate of 11.4% (one-sided permutation P = 0.0017). The flag for possible under-scoring at 50% identified false negatives at 50% in 4 of 8 flagged slides (50.0%; 95% CI 12.5% to 85.8%), compared with a cohort rate of 5.3% (P = 0.0001; Supplementary Section S6).

Referral of the 15% most uncertain slides selected four of the six false-negative slides at the 50% cutoff for review, together with eight of the ten true-positive slides. Sensitivity among auto-reported slides was therefore 2/4 (0.50), compared with 10/16 (0.63) before referral. Supplementary Figure S5 shows how these metrics varied with the percentage of slides referred.

A score-substitution simulation combined model scores for auto-reported slides with a held-out pathologist's recorded scores for referred slides. The simulation's MAE was slightly higher than that of full reader scoring with 10% and 30% of slides referred, and both 95% CIs included zero (Supplementary Section S7).

### 2.5 External Validation

Fusion had the lowest TPS error when development-trained models were applied directly to the external cohort. On the 98 test ROIs, the fused output yielded an MAE of 21.12 percentage points, an ICC of 0.667, and a mean bias of −6.61 percentage points (Table 3). The fused output's MAE was 0.52 percentage points lower than that of density alone (Supplementary Figure S8A).

The external branches showed a different pattern of complementarity from that in the development cohort (Supplementary Figure S8B). For models applied without retraining, detection had lower MAE only in the <1% range, and density had lower MAE in the 1% to <10%, 10% to <50% and ≥50% ranges. The difference arose on E1L3N-stained ROIs, where the detection branch found few or no cells in most ROIs and therefore returned a TPS near 0%. On SP263-stained ROIs, detection again had the lower MAE at a TPS of 50% or above.

| Branch | Condition | ICC | MAE (pp) | r | ρ | Bias (pp) |
|---|---|---|---|---|---|---|
| Fused | Without retraining | 0.667 | 21.12 | 0.685 | 0.674 | −6.61 |
| | Fine-tuned | 0.726 | 19.05 | 0.734 | 0.717 | +1.33 |
| | External-only | 0.718 | 19.18 | 0.723 | 0.721 | +4.19 |
| Density | Without retraining | 0.649 | 21.64 | 0.673 | 0.663 | −7.93 |
| | Fine-tuned | 0.716 | 19.27 | 0.724 | 0.711 | +1.12 |
| | External-only | 0.720 | 19.03 | 0.725 | 0.726 | +3.48 |
| Detection | Without retraining | 0.561 | 23.21 | 0.585 | 0.495 | −12.80 |
| | Fine-tuned | 0.730 | 18.15 | 0.733 | 0.717 | +1.22 |
| | External-only | 0.737 | 17.24 | 0.738 | 0.730 | +2.57 |

**Table 3.** External validation on the IGNITE test cohort (98 ROIs from 21 patients). Rows for application without retraining and fine-tuning pool the 15 models of the three training runs. External-only rows describe one ensemble of five ImageNet-initialized models trained on the external training/validation split. All conditions use the same held-out test ROIs. ICC, intraclass correlation coefficient; MAE, mean absolute error; pp, percentage points; r, Pearson correlation coefficient; ρ, Spearman rank correlation coefficient.

Fine-tuning lowered fused TPS error to 19.05 percentage points and increased ICC to 0.726. The paired MAE difference from application without retraining was −2.07 percentage points (95% CI −7.47 to +2.70; bootstrap clustered by patient). Training only on the external cohort yielded an MAE of 19.18 and an ICC of 0.718. Detection had lower MAE than fusion after both fine-tuning (18.15 percentage points) and external-only training (17.24 percentage points).

Uncertainty ranking also identified higher-error external ROIs when models were applied without retraining. As the referral rate rose from 10% to 50%, mean absolute TPS error decreased from 20.5 to 12.4 percentage points, and sensitivity at the 50% cutoff increased from 0.67 to 0.88 (Figure 5E–F).

The flag analyses adapted count-dependent rules to ROI scale (Supplementary Section S2). Both cutoff-specific flags selected groups with higher rates of their target errors than the external cohort overall (Supplementary Table S3). Precision of the flag for possible over-scoring at 1% for false positives at the 1% cutoff was 58.3%. Precision of the flag for possible under-scoring at 50% for false negatives at the 50% cutoff was 53.3%. The two flags selected 12 and 15 ROIs, respectively. The flag for limited tumor-cell sampling selected 11 ROIs, predominantly resection ROIs.

In the external test set, TPS error was lower on SP263-stained ROIs, the stain used in the development cohort, than on E1L3N-stained ROIs, a clone not seen in training. Before referral, SP263 ROIs had an MAE of 11.06 percentage points and an ICC of 0.878, whereas E1L3N ROIs had an MAE of 29.66 and an ICC of 0.473. With the 15% most uncertain ROIs referred within each clone, sensitivity at the 50% cutoff was 0.92 on SP263 (0.87 before referral). On E1L3N, referral increased the sensitivity of auto-reported ROIs from 0.50 to 0.57, although the MAE gap between staining subsets persisted. Per-clone results and referral curves are given in Supplementary Section S13, Supplementary Table S7 and Supplementary Figure S9.

## 3. Discussion

### 3.1 Model Uncertainty and Reader Disagreement

Slides referred on the basis of model uncertainty also showed greater disagreement among pathologists. Model uncertainty correlated with reader disagreement, and selecting the most uncertain slides for review reduced disagreement among the auto-reported slides (Figure 5D; Supplementary Table S4). This finding gives the uncertainty ranking a clinically relevant role: directing attention to slides that are difficult to score consistently. TPS-Guard uses this ranking to select slides for review and adds cutoff-specific flags that give the reason for review or an advisory on auto-reported slides.

Published automated PD-L1 TPS studies use different reference standards, metrics, and uncertainty-reporting conventions. Kapil et al.[9] evaluated SP263-stained NSCLC needle biopsies against the median of three readers. Pan et al.[14] used a multistage ensemble strategy on SP263- and 22C3-stained NSCLC samples. Van Eekelen et al.[12] assessed an AI system in a multicenter reader study of 100 WSIs, reporting both continuous and categorical TPS agreement. These differences in specimens, references, and agreement measures limit direct numerical comparisons (Supplementary Table S8). Our study relates uncertainty in individual TPS estimates to both automated error and reader disagreement, alongside performance at the clinical cutoffs. Makhlouf et al.[16] described a PD-L1 workflow that directs cases near clinical cutoffs to manual review. TPS-Guard builds on this review-oriented use of automated scoring by combining model uncertainty with branch disagreement, fusion-weight sensitivity, and sampling-related flags. Comparisons with existing referral strategies will require a shared evaluation cohort.

### 3.2 Branch Complementarity and Scene-Aware Fusion

Fusion weights reflect tumor cellularity rather than PD-L1 positivity. In reference-negative slides, incorporating detection counts can increase TPS error relative to density estimation alone. The gain from fusion over equal weighting was small on the development cohort (Supplementary Section S3), and detection alone had lower error after external fine-tuning and external-only training. In the development cohort, detection was more accurate for many slides with high TPS, while fused predictions still tended to underestimate high TPS (Figure 4; Supplementary Figure S1). Disagreement among the branch and alternative fusion estimates can raise the flag for possible under-scoring at 50% when the final TPS remains below 50%.

Macrophage staining can be confused with positive tumor cells. The framework excludes macrophage-rich patches and treats macrophages within retained tumor regions as non-tumor cells. The effect of macrophage burden on fused TPS error was not quantified across the cohort because region-level annotations were available for only the 30 WSIs used to develop tumor segmentation.

### 3.3 Selecting Cases for Pathologist Review

Cutoff-specific flags relate review to the TPS categories used in practice. Near 1% and 50%, a small score difference can change the reported category. The model returns a continuous estimate. A pathologist reports a score near a cutoff as a judgment about which category the slide belongs to, so small differences on the continuous scale are not resolved in the reported score. Disagreement between the model and the pathologists near a cutoff therefore reflects both model error and the way scores are reported, and the two cannot be separated with reader scores alone. TPS-Guard flags these slides for the pathologist with the reason stated rather than deciding the category. The flag for possible over-scoring at 1% identifies a low-positive estimate that may fall below 1%, whereas the flag for possible under-scoring at 50% identifies an estimate below 50% with other model evidence suggesting a score at or above 50%. The flag for limited tumor-cell sampling draws attention to few scoring patches or low predicted tumor cell counts, particularly near a cutoff. These signals give the pathologist specific reasons to recheck the automated estimate alongside the slide.

Uncertainty ranking selects slides for review by identifying disagreement among the 15 models across the TPS range; the flags state the concern at a clinical cutoff, as the reason for referral or as an advisory.

### 3.4 Cross-Center Generalizability

External TPS error was lower for SP263 than for E1L3N, a clone that was absent from development training. Clone, center, and imaging conditions varied together, so their individual contributions remain unresolved.

After adaptation to ROI scale, TPS-Guard identified groups with higher rates of the specified scoring errors in the external cohort. The external rules retained the TPS windows and disagreement criteria while adapting count thresholds and patch conditions (Supplementary Section S2). Substantial continuous TPS error remained.

### 3.5 Limitations and Future Work

The development study was retrospective and conducted at a single center. Reader comparisons had wide bootstrap confidence intervals, and the EBUS analysis involved a small subgroup. External validation comprised 98 ROIs from 21 patients and omitted tumor segmentation. Sex-stratified performance, reader confidence, and clinical outcomes after review were not assessed. Positive tumor cells made up 6% of the annotated nuclei, which left the training data imbalanced between PD-L1-positive and PD-L1-negative tumor cells, the two classes on which TPS depends. More annotated positive tumor cells will be needed to address this imbalance in the cell annotations; in the development cohort, retraining with the positive class reweighted did not improve agreement with the pathologists (Supplementary Section S15).

Flag criteria were developed on the cohort used for evaluation, and external count quantiles were calculated from the external test cohort's model outputs. Prospective studies should use fixed, prespecified rules on new whole slides and compare pathologist scoring with and without TPS-Guard cues, including accuracy and review time.

## 4. Conclusion

In this retrospective study, model uncertainty identified slides with larger TPS errors and greater disagreement among pathologists. TPS-Guard added cutoff-specific flags that enriched defined error events and recorded a reason for review. These findings support uncertainty-guided triage of automated TPS estimates. External ROI analyses supported this selection behavior after adaptation of count-dependent rules, despite substantial residual scoring error. The next step is prospective evaluation of how these cues affect pathologist scoring and review time.

## Acknowledgments

The authors thank all contributors to data collection and annotation.

## Data Availability Statement

The images and associated annotations from the UHCW development cohort are not publicly available because they are governed by the REC-approved protocol and institutional data-sharing restrictions. Access requests may be directed to the corresponding author and will be considered subject to REC, institutional governance, and data-use approvals. The external validation dataset (IGNITE) is publicly available via Zenodo (version 1, https://doi.org/10.5281/zenodo.15674785)[26] and is described by Spronck et al.[20] The original training code, complete experiment configuration snapshots, and inference and evaluation code will be made publicly available upon acceptance of the manuscript.

## Declaration of Competing Interests

Declarations of interest: David Snead is a shareholder and Director of Histofy Ltd. The other authors declare no competing interests.

## Ethics Approval and Consent to Participate

Ethics approval was granted by the North West – Lancaster Research Ethics Committee for "Digital pathology algorithm development" (REC reference 15/NW/0843; Integrated Research Application System (IRAS) project ID 189095; protocol number 1; favorable opinion dated 16 October 2015). The study was conducted in accordance with the Declaration of Helsinki and the REC-approved application, protocol and supporting documentation, with the required host-organization management permission/R&D approval. Individual informed consent was not required for this analysis because only de-identified archival tissue images and associated data were used under the approved protocol.

## Funding Statement

This work was supported by PhD funding provided to Kesi Xu by the Department of Computer Science, University of Warwick. The funder had no role in study design; data collection, analysis, or interpretation; manuscript preparation; or the decision to submit the article.

## Author Contributions

Kesi Xu: Conceptualization, Methodology, Software, Validation, Formal analysis, Investigation, Data curation, Visualization, Project administration, Writing – original draft, Writing – review & editing. Ayesha Azam: Data curation, Investigation, Resources, Writing – review & editing. Tanvier Kaur: Investigation, Writing – review & editing. Monika Kondratowicz: Data curation, Writing – review & editing. Kinza Asim: Data curation, Writing – review & editing. Brinder Chohan: Data curation, Writing – review & editing. Paapa Appiah-Odame: Data curation, Writing – review & editing. Zhao Chen: Methodology, Writing – review & editing. David Snead: Conceptualization, Resources, Investigation, Supervision, Writing – review & editing. Nasir Rajpoot: Conceptualization, Methodology, Supervision, Writing – review & editing. All authors reviewed and revised the manuscript, approved the final version, and agree to be accountable for all aspects of the work.

## Declaration of Generative AI and AI-Assisted Technologies in the Manuscript Preparation Process

During the preparation of this work, the authors used Claude (Anthropic) in order to perform basic spelling, grammar, and punctuation checks. After using this tool, the authors reviewed and edited the content as needed and take full responsibility for the content of the published article.

# Automated PD-L1 Tumor Proportion Scoring with Uncertainty-Guided Triage in Non-Small Cell Lung Cancer

Kesi Xu[1,*], Ayesha Azam[2], Tanvier Kaur[1], Monika Kondratowicz[3], Kinza Asim[2], Brinder Chohan[4], Paapa Appiah-Odame[5], Zhao Chen[6], David Snead[1,2], Nasir Rajpoot[1]

[1] Tissue Image Analytics Centre, Department of Computer Science, University of Warwick, Coventry CV4 7AL, UK

[2] Department of Pathology, University Hospitals Coventry and Warwickshire NHS Trust, Clifford Bridge Road, Coventry CV2 2DX, UK

[3] Department of Pathology, University Hospitals Birmingham NHS Foundation Trust, Mindelsohn Way, Edgbaston, Birmingham B15 2GW, UK

[4] Department of Histopathology, Royal Derby Hospital, Derby, UK

[5] University Hospitals Birmingham NHS Foundation Trust, Birmingham, UK

[6] School of Information and Intelligent Science, Donghua University, Shanghai, China

***Corresponding author:** Kesi Xu (kesi.xu@warwick.ac.uk)

## Supplementary Material

Section S1 reports inter-pathologist agreement diagnostics, including how range reports were encoded, and the main results with the original reader scores. Section S2 specifies the TPS-Guard rules and adaptations for external regions of interest (ROIs). Section S3 describes architecture, fusion, and training, Section S4 shows tumor segmentation examples, and Section S5 describes annotations. Section S6 reports sensitivity analyses within the development cohort and flag-enrichment statistics, and Section S7 presents matched-slide comparisons and a score-substitution simulation. Sections S8–S10 cover development-cohort diagnostics for residuals and referral orderings, specimen-stratified agreement and uncertainty. Section S11 illustrates cell counting in an external ROI, Section S12 summarizes external validation, and Section S13 reports external clone-subgroup analyses. Section S14 places the study in context with selected automated programmed death-ligand 1 (PD-L1) tumor proportion score (TPS) systems. Section S15 reports a class-reweighting experiment. Superscript numbers refer to the main-text reference list.

### S1. Inter-Pathologist Agreement Diagnostics

For analysis, reports of $< 1\%$ and $> 1\%$ were encoded as 0.5% and 1.5%, respectively. The two reports of 1–5% were encoded as 3%, and numerical scores retained their recorded precision. In the Bland–Altman analysis (Supplementary Figure S1I), limits of agreement were the mean paired difference $\pm 1.96$ times its population standard deviation (SD).

Supplementary Table S1 gives the main results with the original reader scores, before three of P1's scores were corrected (main text Section 1.1.1).

### S2. TPS-Guard Rule Specification

TPS-Guard thresholds were derived during development, so the development-cohort analyses remain internal evaluations. The rules below were implemented for the reported analyses. TPS values are percentages, and counts are continuous model estimates. Flags are assigned without using reference TPS or reader scores.

### S2.1 Inputs and Patch-Derived Features

Let $\mathrm{F}$ denote the final fused TPS, $\mathrm{T}_1, \ldots, \mathrm{T}_{15}$ the fused TPS values from the 15 models, $\mathrm{D}$ the density-branch TPS (aggregation specified below), $\mathrm{C}^{+}$ the predicted positive tumor cell count, $\mathrm{N}$ the total predicted tumor cell count, and P the number of scoring patches. For development WSIs, $\mathrm{C}^{+}$ and N are detection-branch counts averaged across the 15 models. P is the integer part of the mean number of scored patches across models. The ranking score is the population standard deviation of $\mathrm{T}_1, \ldots, \mathrm{T}_{15}$. The final TPS uses the three held-out models or count averaging across 15 models, as specified in main text Section 1.7. Uncertainty and patch-derived features use the 15-model outputs. For development WSIs, each model's TPS is $\mathrm{T_k} = \frac{100\mathrm{U_k}}{\mathrm{U_k} + \mathrm{V_k} + 10^{-6}}$, where $\mathrm{U_k}$ and $\mathrm{V_k}$ are fused positive and negative count sums for model $k$.

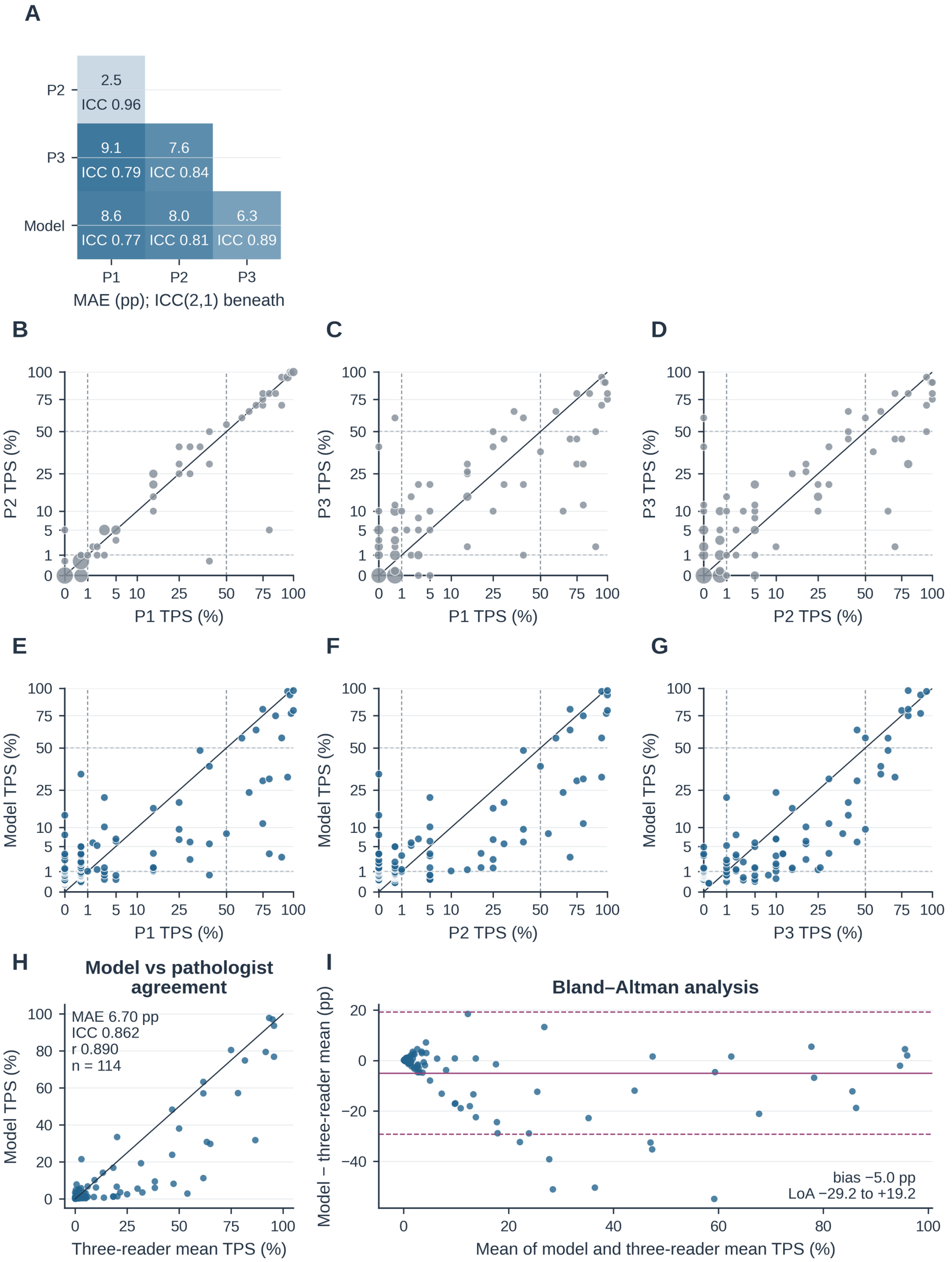

**Supplementary Figure S1.** Pairwise agreement between pathologists and between the model and each pathologist on the 114 whole-slide images (WSIs) without cell-level annotations. (A) MAE and ICC(2,1) for every pair. (B–D) Pathologist pairs; (E–G) model versus each pathologist. Each scatter in B–G shows one point per slide, with the line of identity and the 1% and 50% cutoffs; larger points mark overlapping slides, and axes use square-root scales. (H) D2Count-Net versus three-reader mean TPS. (I) Bland–Altman analysis of D2Count-Net against the three-reader mean; the solid line is the mean bias and the dashed lines are the 95% limits of agreement. ICC, intraclass correlation coefficient; MAE, mean absolute error; r, Pearson correlation coefficient; LoA, limits of agreement; pp, percentage points.

| Quantity | Original scores | Corrected scores (main analysis) |
|---|---|---|
| *Before referral (114 WSIs)* | | |
| Model MAE vs three-reader mean (pp) | 6.85 | 6.70 |
| Pairwise pathologist MAE (pp) | 7.54 | 6.41 |
| Leave-one-reader-out MAE, model / single pathologist (pp) | 7.07 / 7.00 | 6.96 / 6.01 |
| Spearman ρ, model uncertainty vs absolute error | 0.747 | 0.766 |
| Spearman ρ, model uncertainty vs reader SD | 0.682 | 0.714 |
| *After referral of the 15% most uncertain slides (97 auto-reported)* | | |
| Model MAE vs three-reader mean (pp) | 5.23 | 4.96 |
| Pairwise pathologist MAE (pp) | 5.48 | 4.93 |
| Leave-one-reader-out MAE, model / single pathologist (pp) | 5.45 / 5.16 | 5.18 / 4.61 |

**Supplementary Table S1.** Main results with the original and the corrected reader scores on the 114 WSIs without cell-level annotations. Three of P1's scores (WSIs 152, 371 and 400) were corrected by P1 after a check of slides with large differences between readers; all other scores are the original independent scores. The main text uses the corrected scores. MAE, mean absolute error; pp, percentage points; SD, standard deviation of the three reader scores.

**Final TPS calculation.** Development WSIs use the three held-out models for the 40 cell-annotated slides and count averaging across 15 models for the other 114 slides, as described in main text Section 1.7. Final TPS is the fused positive tumor cell count divided by the total fused tumor cell count, limited to 0–100%, and is 0% when no tumor cells are predicted. External ROIs use the same calculation.

For development WSIs, D is computed from density-branch count sums using the same model assignment as the final fused TPS: the three held-out models for the 40 cell-annotated WSIs and count averaging across 15 models for the other 114 WSIs. For external ROIs, density counts are summed within each model and averaged across the 15 models before TPS calculation.

Positive-cell concentration K is computed within each model as the proportion of predicted fused-positive cells in the five patches with the largest positive counts (all patches if fewer than five are available), then averaged over models with predicted positive cells.

High-TPS patch burden B is the total predicted tumor cell count in patches with patch TPS $\geq 70\%$, divided by the total tumor cell count across all scored patches and models. B weights patches by cell count.

Two sensitivity estimates, $A_{0.5}$ and $A_{1.0}$, are obtained by replacing the scene-conditioned detection weight with a constant of 0.5 or 1.0, respectively, in every patch. The resulting TPS values are averaged across models.

### S2.2 Binary Flags on Development WSIs

**Possible over-scoring at 1%:** $1 \leq F < 5$ AND at least one of: (i) $\min(T_1, \ldots, T_{15}) < 1$ and $\max(T_1, \ldots, T_{15}) \geq 1$; (ii) $\frac{F}{1.2} < 1$; (iii) $C^{+} < 50$; or (iv) $K \geq 0.7$. The $\frac{F}{1.2}$ criterion rescales the TPS numerically; it does not rerun cell detection with a different staining or detection threshold.

**Possible under-scoring at 50%:** $20 \leq F < 50$ AND at least one of: (i) the 90th percentile of $T_1, \ldots, T_{15}$ is $\geq 50$; (ii) $\max(T_1, \ldots, T_{15}) \geq 50$; (iii) $D \geq 50$; (iv) $A_{0.5} \geq 50$ or $A_{1.0} \geq 50$; or (v) $B \geq 0.10$.

**Limited tumor-cell sampling:** at least one of (i) $N < 1{,}000$; (ii) $P < 20$; (iii) the specimen type is endobronchial ultrasound (EBUS)-guided or other biopsy, $N < 3{,}000$, and either $F < 5$ or $35 \leq F < 65$; or (iv) a count-based interval crossing 1% or 50%. For (iv), $p = \mathrm{clip}(\frac{F}{100}, 0, 1)$, and $n_{eff}$ equals N if N is finite and positive, otherwise 1. The lower and upper endpoints are:

$$(L, U) = 100 \times [p \pm 1.96 \sqrt{\frac{p(1-p)}{n_{eff}}}]$$

A crossing means $L < c \leq U$ for $c = 1$ or $c = 50$. This Wald-type interval is a warning signal based on predicted cell counts; it is not a confidence interval for the reference TPS.

For referral, at target retention $r$ in a cohort of $n$ cases, cases are ranked by increasing model SD. The first $\lceil nr \rceil$ cases are auto-reported, rounding the target count up to the nearest integer, and the others are referred for pathologist review. In the 114 WSIs without cell-level annotations, the flags for possible over-scoring at 1%, possible under-scoring at 50% and limited tumor-cell sampling identify 21, 8, and 26 WSIs, respectively. Their union contains 45 WSIs. Retention is the proportion of cases auto-reported, that is, 1 minus the referral rate used in the main text.

### S2.3 Adaptations for External ROIs

The external analysis uses fused positive and negative counts to define $C^+$ and $N$. TPS from each model is limited to $[0, 100]$ before calculating model uncertainty. For the 98 ROIs, the $C^+ < 50$ condition becomes $C^+ < Q_{0.10}(C^+)$, $N < 1{,}000$ becomes $N < Q_{0.10}(N)$, and the small-specimen condition requires a tissue microarray specimen, $N < Q_{0.30}(N)$, and the same cutoff-proximity condition. Quantiles use linear interpolation. $P < 20$ becomes $P < 3$. For the clone-specific analyses in Section S13, count quantiles and uncertainty rankings are recalculated within each clone. The count-based interval criterion is not used. The TPS windows, cutoff-crossing tests, $\frac{F}{1.2}$ criterion, concentration threshold 0.7, and high-TPS burden threshold 0.10 are unchanged.

External patch features use the same five-patch concentration, $\geq 70\%$ patch-TPS threshold, and fixed weights 0.5 and 1.0.

For the 15-model ensemble applied to the external cohort without retraining, the 10th percentiles of the predicted positive and total tumor cell counts were 0.0263 and 0.113, and the 30th percentile of the total count was 0.977. These thresholds are relative to the model outputs in this cohort and are not clinical minimum cell counts.

All flag inputs were available for the 154 development WSIs and the 98 external ROIs.

## S3. Adaptive Fusion, Architecture, and Training Details

The dual-branch cell counting network (D2Count-Net) uses a shared EfficientNet-B0[22] image encoder pretrained on ImageNet and two independent U-Net-style decoders with skip connections.[23] The detection decoder produces a 3-channel output (PD-L1-positive tumor cells, PD-L1-negative tumor cells, and non-tumor cells) via sigmoid activation. During training, counts are sums of soft peak responses (Section S3.2). For the reported results, each class count is the sum of p over local maxima with $p \geq 0.5$, so each retained peak counts as p cells. The density decoder produces a 2-channel density map using a learned per-channel scale and bias, followed by a softplus activation and a fixed output scaling factor of 0.002. Density counts are spatial sums of the map. A tissue scene classifier predicts one of four cellularity categories from encoder features after global average pooling. It uses a two-layer multilayer perceptron (MLP; $320 \rightarrow 64 \rightarrow 4$), rectified linear unit (ReLU) activation and 30% dropout. The categories are low cellularity (1–14 tumor cells per patch), medium (15–39), high ($\geq 40$) and background (zero tumor cells).

### S3.1 Scene-Based Fusion Weights

The scene probabilities and class-specific priors give the weighted sum:

$$a = \Sigma \, p_c \cdot \alpha_c^{prior} \quad \text{(S1)}$$

where $p_c$ is the predicted probability of scene class c, and $\alpha_c^{prior}$ is the fixed detection-branch prior for that class. The priors are 0.7 for low cellularity, 0.5 for medium cellularity, 0.3 for high cellularity and 0.5 for background. The weighted sum a from Equation (S1) is converted to the detection weight by:

$$\alpha = 0.01 + 0.98 \cdot \text{sigmoid}[6(a - 0.5)]$$

This mapping bounds $\alpha$ to $[0.01, 0.99]$; with the fixed priors, $\alpha$ lies between 0.24 and 0.76. The priors and mapping are fixed, while the detection weight varies with each patch's scene probabilities. The density weight is $1 - \alpha$. The fused cell counts are:

$$\hat{y}_{fused} = \alpha \cdot \hat{y}_{det} + (1 - \alpha) \cdot \hat{y}_{dens} \quad \text{(S2)}$$

### Fixed Priors and Constant-Weight Check

The four priors were fixed across the development and external cohorts, with no cohort-specific weight search for the reported results. In a post hoc comparison, we replaced the scene-conditioned weight with a constant $\alpha = 0.5$ on the same branch outputs. All other steps were unchanged, including the three held-out models for the 40 cell-annotated WSIs and the averaging of cell counts across the 15 models for the other 114 WSIs. On the 114 WSIs without cell-level annotations, mean absolute error (MAE) rose from 6.70 to 6.77 percentage points and intraclass correlation coefficient (ICC) fell from 0.862 to 0.857.

No additional post hoc threshold on tumor cell count was applied to the continuous TPS estimate after fusion. TPS-Guard uses limited tumor-cell sampling as a flag, not a referral criterion, and leaves the automatically generated TPS value unchanged.

### S3.2 Architecture and Training Details

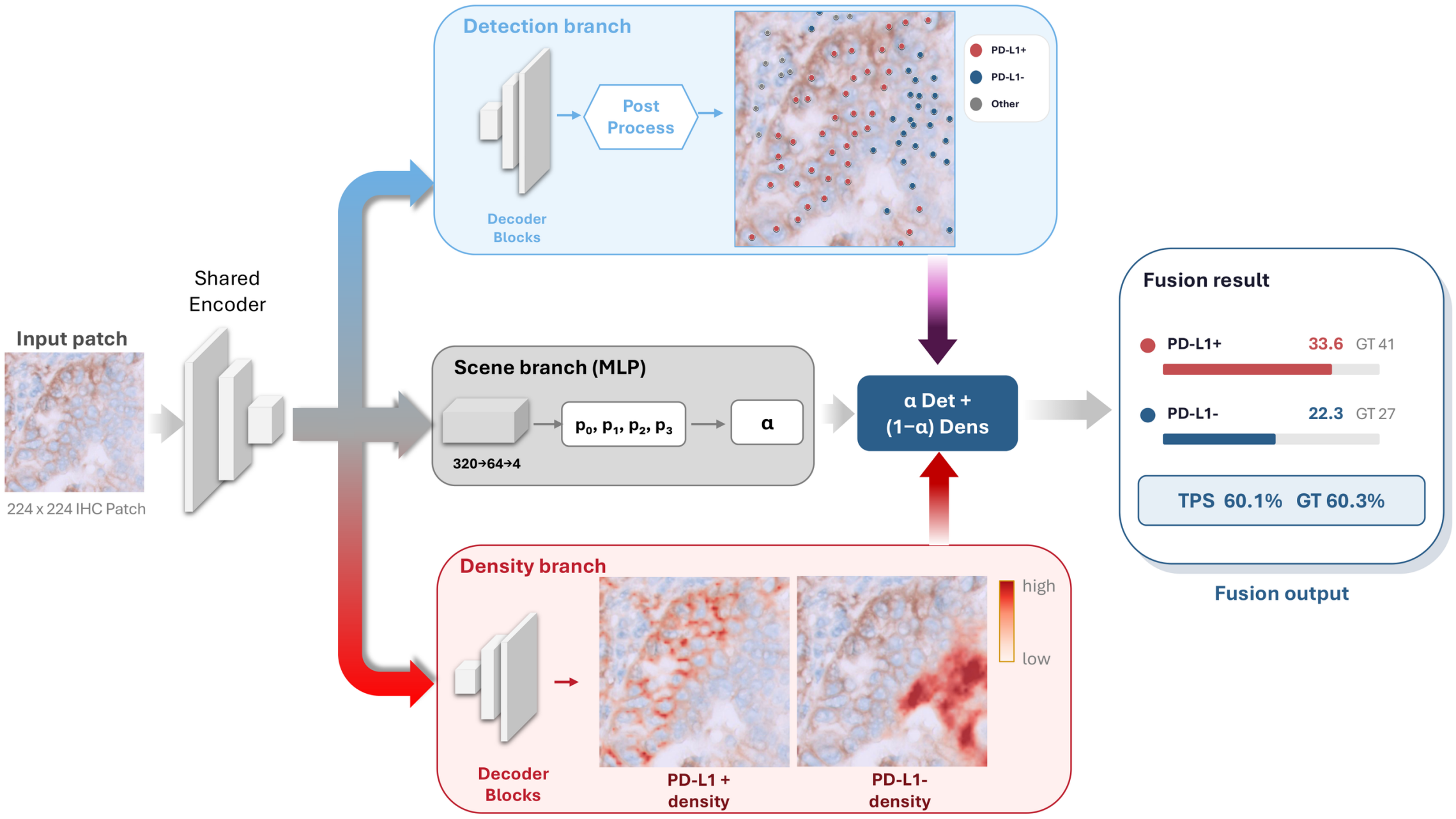

**Supplementary Figure S2.** D2Count-Net architecture for cell-level analysis. A shared, pretrained encoder feeds three parallel branches. The detection branch (top) marks individual cells with class labels (PD-L1-positive tumor, PD-L1-negative tumor, non-tumor). In the scene branch (middle), a $320 \rightarrow 64 \rightarrow 4$ multilayer perceptron (MLP) followed by softmax outputs probabilities $p_0$, $p_1$, $p_2$, and $p_3$ for low, medium, high, and background cellularity, respectively. These probabilities weight the fixed scene priors to give a (Equation S1), which is converted to the detection weight α by the sigmoid mapping in Section S3.1. The density branch (bottom) regresses continuous PD-L1-positive and PD-L1-negative density maps. The fusion block applies α to detection counts (Det) and $1 - \alpha$ to integrated density counts (Dens), separately for positive and negative tumor cells (Equation S2). The example output (right) compares predicted counts with ground truth (GT) for a representative patch.

### Tumor Segmentation Model

The tumor segmentation network uses the Virchow2 foundation model[21] as a frozen feature extractor. Input PD-L1 WSIs are divided into non-overlapping $224 \times 224$-pixel patches at $20\times$ magnification. A linear classifier on the class token features assigns each patch to tumor, non-tumor, or macrophage-rich tissue. Macrophage-rich patches are merged into non-tumor at inference. Tumor regions smaller than about 0.07 $mm^2$ are then removed, and the tumor map is intersected with the tissue mask to define the region analyzed in Stage 2.

**Scene-Aware D2Count-Net Cell Analysis Model**

**Shared encoder.** An EfficientNet-B0[22] pretrained on ImageNet is the shared feature extractor. Encoder weights are frozen for the first 5 epochs, then fine-tuned jointly with the rest of the network.

**Dual decoders.** Two independent U-Net-style decoders[23] with skip connections upsample encoder features to one quarter of the input resolution, and the output maps are bilinearly resized to the input size. Each decoder is supervised by its branch-specific map and count losses.

**Detection branch.** Produces a 3-channel output (PD-L1-positive tumor cells, PD-L1-negative tumor cells, and non-tumor cells) via sigmoid activation. For each channel, $p(x, y)$ is the sigmoid probability on an $H \times W$ map. A small coordinate perturbation is added to locate local maxima:

$$\tilde{p}(x,y) = p(x,y) + 10^{-4}[\frac{0.1y}{H-1} + \frac{0.01x}{W-1}]$$

The local-maximum indicator $m(x, y)$ is one where $\tilde{p}$ equals its $5 \times 5$ max-pooled value (stride 1, padding 2), and zero elsewhere. The soft response is:

$$s(x,y) = p(x,y)m(x,y)\sigma[10(p(x,y) - 0.5)]$$

Here, σ is the sigmoid function, and the channel count is $\sum_{x,y} s(x, y)$. The coordinate perturbation breaks ties and is used only to locate maxima. This soft count is used for training and model selection; for the reported results, the channel count is the sum of $p(x, y)$ over local maxima with $p(x, y) \geq 0.5$.

**Density branch.** Produces a 2-channel density map using a learned per-channel scale and bias, followed by softplus activation and a fixed output scaling factor of 0.002. Counts are spatial sums of the map.

**Scene classifier.** A two-layer classifier (320 → 64 → 4, ReLU activation, 30% dropout) uses encoder features after global average pooling to output posterior probabilities for background (0 tumor cells), low (1–14), medium (15–39), and high (≥ 40) cellularity.

**Scene-based fusion.** The posterior-weighted sum of the four class-specific priors in Equation (S1) gives the intermediate value a. The fixed priors and sigmoid mapping in Section S3.1 convert this value to the detection weight α. Detection receives the larger weight in sparse patches, and density estimation receives the larger weight in crowded patches.

### Training Protocol

D2Count-Net was trained for up to 150 epochs per fold using AdamW optimization (initial learning rate $2 \times 10^{-4}$, cosine annealing to $1 \times 10^{-6}$), batch size 48, gradient clipping at a maximum norm of 5.0, and automatic mixed-precision training. Augmentation used hue, saturation and value shifts, red, green and blue (RGB) channel shifts, random brightness and contrast changes, vertical flips, and random 90° rotations. The EfficientNet-B0 encoder was frozen for the first 5 epochs and then fine-tuned jointly. Validation MAE was monitored for early stopping, with a patience of 25 epochs. Across the three training runs (15 cross-validation models), training stopped at 80–150 epochs (three of the 15 models reached the 150-epoch limit), and the selected models had validation MAE of 2.13–2.88 cells per class per patch (mean 2.57). For the 40 cell-annotated WSIs in the analysis cohort, the final TPS estimate used only the three held-out models. For the other 114 WSIs, cell counts were averaged across the 15 models (main text Section 1.7).

**External fine-tuning.** Each development-trained model was fine-tuned using the same multi-task loss, with a learning rate of $1 \times 10^{-4}$ and a maximum of 100 epochs. Early stopping used a patience of 20 epochs, and the encoder was frozen for the first 5 epochs. The 15 models used the same external training/validation partition and differed in their development-cohort initialization. Their cell counts were averaged across models.

External-only training. Five models were trained independently from ImageNet-pretrained encoder weights, using a learning rate of $2 \times 10^{-4}$, a maximum of 150 epochs, and early stopping with a patience of 25 epochs. The architecture, external training/validation split, and held-out test ROIs were the same as for fine-tuning. Cell counts were averaged across the five models.

**Random seeds.** The three independent development-cohort training runs used random seeds 42, 123 and 456, respectively, with five cross-validation models per run. External-only training used five models with random seeds 42, 43, 44, 45 and 46. The paired patient-level bootstrap comparing external application without retraining with fine-tuning used 10,000 resamples.

Agreement, correlation and referral statistics were calculated in Python 3.11.15 with NumPy 2.4.3, pandas 3.0.1 and SciPy 1.17.1.

The training objective comprises nine patch-level loss components. All detection, density, other-cell and fused count losses use mean normalized absolute error, $\mathrm{mean}[\frac{|\hat{c} - c|}{c + 1}]$, over the corresponding samples and count channels, where $\hat{c}$ and $c$ are the predicted and annotated counts, respectively. The detection-map loss is an equally weighted sum of focal-weighted binary cross-entropy and Dice loss. The density-map loss is mean squared error; the peak-preservation loss is mean squared error at the 100 spatial locations with the largest target density values in each sample and density channel. Scene classification uses weighted cross-entropy with weights proportional to (3.85, 2.51, 6.45, 5.35) for low, medium, high and background scenes, normalized to sum to four.

For the detection-map loss, let z denote logits, $q = \sigma(z)$, g the target map clipped to $[0, 1]$, and $b = 1[g \geq 0.3]$. Binary cross-entropy (BCE) is evaluated directly from the logits. The focal term is the mean of $(1 - p_t)^2 \mathrm{BCE}(z, g)$, with $p_t = qg + (1 - q)(1 - g)$. The Dice term is 1 minus the sample/channel mean of $\frac{2\sum qb + \varepsilon}{\sum q + \sum b + \varepsilon}$, where spatial locations are summed and $\varepsilon = 10^{-6}$. The detection-map loss is 0.5 times each term.

For sample i, let $d_i$ be the mean absolute difference between detection and density counts across the positive and negative tumor cell channels. The fused-error margin loss is the sample/channel mean of $\max(|\hat{c}_{\mathrm{fused}} - c| - d_i, 0)$. This term penalizes fused count error above the between-branch count difference.

The detection, density and shared task totals are:

$$L_{\mathrm{det}} = L_{\mathrm{det,map}} + L_{\mathrm{det,count}} + L_{\mathrm{other,count}}$$

$$L_{\mathrm{dens}} = 0.5 L_{\mathrm{dens,map}} + 3 L_{\mathrm{dens,count}} + L_{\mathrm{peak}}$$

$$L_{\mathrm{shared}} = 2 L_{\mathrm{scene}} + L_{\mathrm{fused}} + 0.1 L_{\mathrm{margin}}$$

The network objective is:

$$w_{\mathrm{det}} L_{\mathrm{det}} + w_{\mathrm{dens}} L_{\mathrm{dens}} + L_{\mathrm{shared}}$$

Dynamic weights apply to the two branch totals, while the shared total is added separately.

The relative contributions of the detection and density tasks are adjusted during training using $w = \mathrm{softmax}(u)$, with both elements of u initialized to zero so that the initial task weights are equal. When task reweighting begins, the reference task losses are stored without changing the weights. For each subsequent update, the ratio

$$r_j = \frac{L_{j,current}}{L_{j,previous} + 10^{-8}}$$

compares the current and preceding batch losses for task $j$. These loss values are held constant during the weight update. Before updating the network parameters, Adam (learning rate 0.025) updates u to minimize:

$$\sum_j w_j r_j + 10^{-5} \sum_j w_j \log(w_j + 10^{-8})$$

The updated weights are then used in the network objective defined above.

**Ensemble Uncertainty Computation**

Model uncertainty was measured by the standard deviation of TPS estimates from the 15 cross-validation models (main text Section 1.8). Its association with scoring error and inter-reader disagreement is reported in Section S10.

## S4. Tumor Segmentation Examples

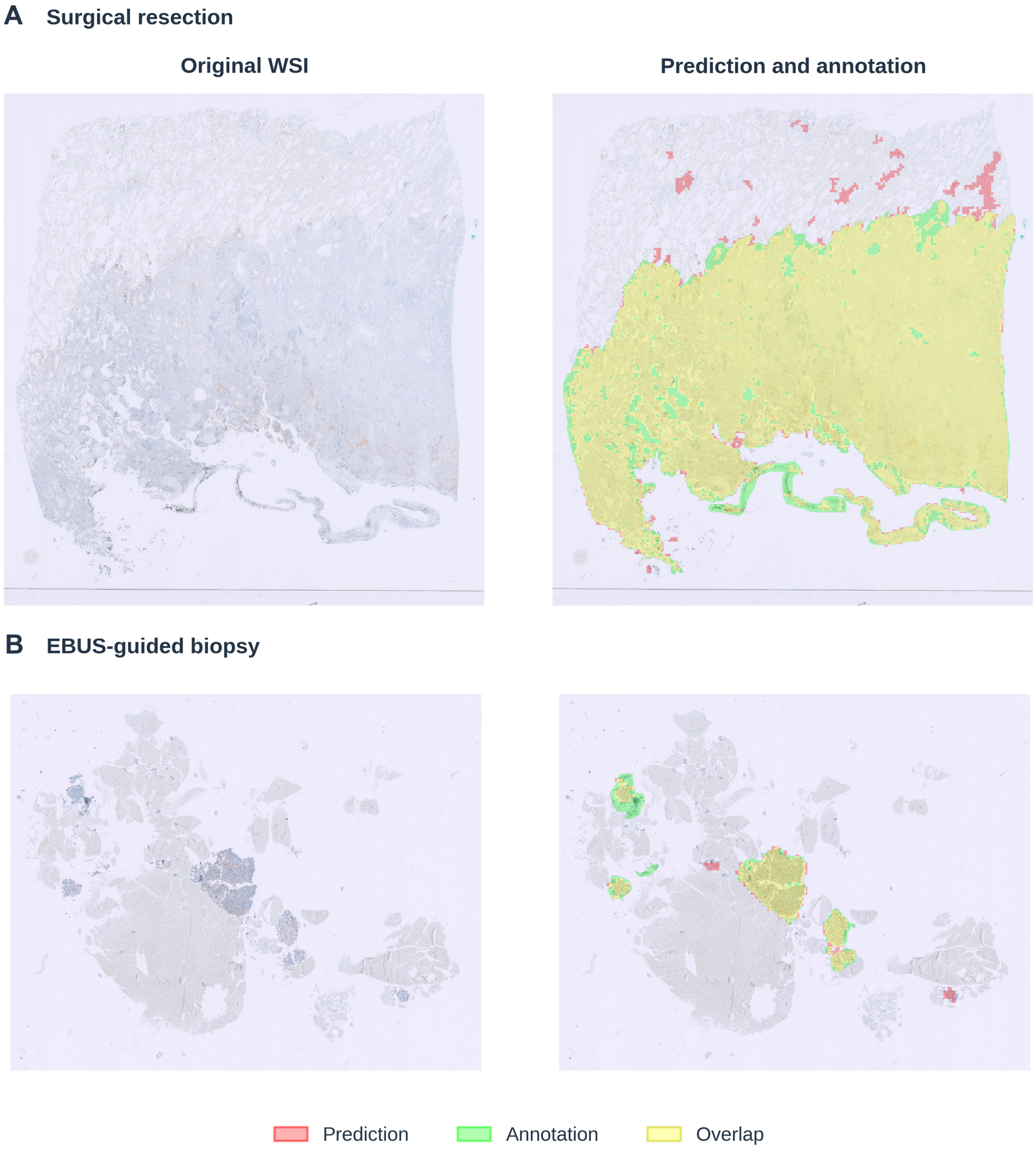

**Supplementary Figure S3.** Representative automated tumor segmentation. (A) Surgical resection specimen. (B) EBUS-guided biopsy. Red, algorithm-predicted tumor; green, pathologist-annotated ground truth; yellow, concordant regions.

### S5. Annotated Cell-Type Distribution

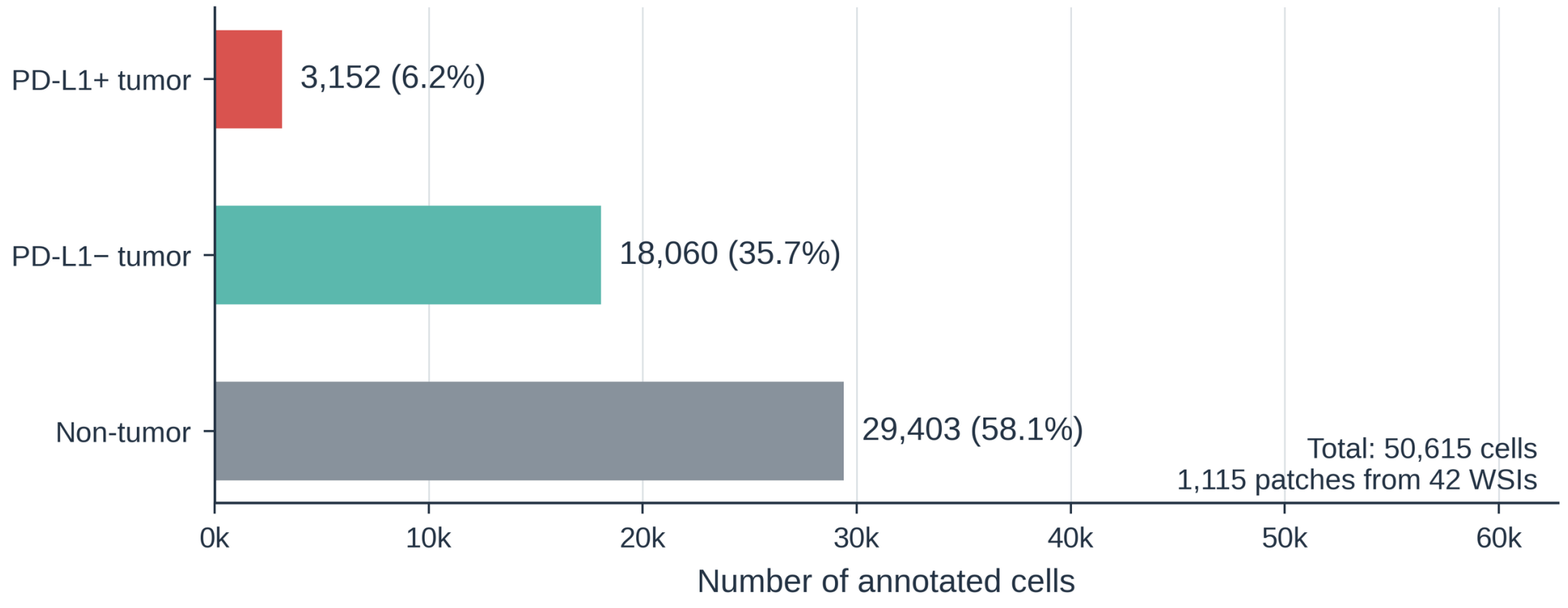

**Supplementary Figure S4.** Distribution of annotated cell types across the 1,115 annotated patches from 42 WSIs.

### S6. TPS-Guard Development-Cohort Sensitivity Analyses and Flag Enrichment

On the 114 WSIs without cell-level annotations, we compared uncertainty ranking with an alternative ordering that refers flagged slides first (flags first). Both orderings used the fixed priors and mapping in Section S3.1 and unchanged TPS-Guard rules. Because TPS-Guard rules were developed using the development cohort, these remain internal sensitivity analyses and their performance estimates may be optimistic.

| **Configuration** | **Actual retention** | **MAE (pp)** | **ICC** | **Specificity at 1%** | **Sensitivity at 50%** | **Three-tier discordance** |
|---|---|---|---|---|---|---|
| Uncertainty ranking (main analysis) | 0.851 | 4.96 | 0.762 | 0.755 | 0.500 | 0.268 |
| Flags first | 0.851 | 4.73 | 0.908 | 0.784 | 0.875 | 0.237 |
| No referral | 1.000 | 6.70 | 0.862 | 0.755 | 0.625 | 0.263 |

**Supplementary Table S2.** TPS-Guard performance on the 114 WSIs without cell-level annotations at 85% target retention. Uncertainty ranking (main analysis) refers slides by model uncertainty alone; flags do not change referral. MAE, mean absolute error; pp, percentage points; ICC, intraclass correlation coefficient. Three-tier agreement is the three-tier accuracy of the main text, and three-tier discordance is 1 minus it.

On the 114 WSIs without cell-level annotations, 97 slides were auto-reported at the 85% target retention (15% referred). Flags first gave lower MAE than uncertainty ranking (4.73 vs 4.96 percentage points). Flags first also raised ICC from 0.762 to 0.908 and sensitivity at the 50% cutoff from 2/4 to 7/8.

| Cohort | Flag | Precision | 95% CI | Base rate | One-sided P |
|---|---|---|---|---|---|
| UHCW | Possible over-scoring at 1% | 33.3% (7/21) | 14.3–55.2% | 11.4% | 0.0017 |
| UHCW | Possible under-scoring at 50% | 50.0% (4/8) | 12.5–85.8% | 5.3% | 0.0001 |
| External (15 models) | Possible over-scoring at 1% | 58.3% (7/12) | 33.3–83.3% | 22.4% | — |
| External (15 models) | Possible under-scoring at 50% | 53.3% (8/15) | 28.6–75.0% | 14.3% | — |

**Supplementary Table S3.** TPS-Guard flag-enrichment analysis. Precision is the proportion of flagged cases meeting the specified target event: a false positive at the 1% TPS cutoff for the flag for possible over-scoring at 1%, and a false negative at the 50% TPS cutoff for the flag for possible under-scoring at 50%. Development-cohort intervals use 2,000 WSI bootstrap samples on the 114 WSIs without cell-level annotations. External intervals use 2,000 patient-cluster bootstrap samples, retaining all ROIs from each of the 21 sampled patients. Observed flag assignments and target events are held fixed during resampling. Development-cohort one-sided *P* values use 10,000 permutations of flag labels across WSIs. External precision and intervals are descriptive and are reported without *P* values. CI, confidence interval; UHCW, University Hospitals Coventry and Warwickshire.

For the development cohort, each base rate is the cohort-level prevalence of the corresponding target event. The external base rate was calculated from the 15-model ensemble because cutoff errors depend on model predictions.

## S7. Matched-Slide Comparisons and a Score-Substitution Simulation

We recomputed pairwise inter-pathologist MAE and ICC on the same auto-reported slides at each retention level to examine whether the model-derived ranking selects cases with greater reader disagreement. Model scores were compared with the three-reader mean. A separate analysis simulated score substitution under uncertainty ranking, with no flags. Auto-reported cases took the model score, and referred cases took a held-out pathologist's score. This combined score and the same pathologist's scores for all slides were compared with the mean of the other two readers, with each reader held out in turn. The simulation used recorded scores, without asking pathologists to reassess slides using the review cues. For this analysis, the auto-reported case count was rounded upward, as in the referral rule in Section S2. This gives 109, 103, 97, 92, 86 and 80 auto-reported slides at 95%, 90%, 85%, 80%, 75% and 70% target retention, respectively.

| Target retention | Auto-reported n | Model MAE | Pathologist MAE, same slides | Model ICC | Pathologist ICC, same slides | Score-substitution MAE, all slides | Pathologist MAE, all slides |
|---|---|---|---|---|---|---|---|
| 100% | 114 | 6.70 | 6.41 | 0.862 | 0.862 | 6.96 | 6.01 |
| 95% | 109 | 5.94 | 5.72 | 0.876 | 0.869 | 6.83 | 6.01 |
| 90% | 103 | 5.48 | 5.28 | 0.857 | 0.853 | 6.69 | 6.01 |
| 85% | 97 | 4.96 | 4.93 | 0.762 | 0.760 | 6.49 | 6.01 |
| 80% | 92 | 4.92 | 4.17 | 0.758 | 0.804 | 6.95 | 6.01 |
| 75% | 86 | 3.63 | 3.65 | 0.847 | 0.799 | 6.28 | 6.01 |
| 70% | 80 | 2.45 | 2.15 | 0.941 | 0.941 | 6.37 | 6.01 |

**Supplementary Table S4.** Model and pathologist error on the 114 WSIs without cell-level annotations as uncertain slides are referred. Slides are ranked by model uncertainty. Columns 3–6 compare the model with the three-reader mean and pathologists with each other on the same auto-reported slides. Columns 7–8 give a score-substitution simulation with each pathologist held out in turn: referred slides take that pathologist's score, and all scores are compared with the mean of the other two pathologists. MAE, TPS percentage points; ICC, ICC(2,1).

Pairwise pathologist MAE fell from 6.41 on all 114 slides to 5.28 on the 103 slides auto-reported at 90% retention, consistent with greater reader disagreement among the cases referred by the model-derived ranking. On the same auto-reported slides, model MAE was within 0.75 percentage points of inter-pathologist MAE at every retention level (largest gap at 80%: 4.92 versus 4.17). Model ICC was within 0.05 of the matched inter-pathologist ICC at every retention level (0.758 versus 0.804 at 80% and 0.847 versus 0.799 at 75%). Against the mean of the other two readers, the model's error was 0.36 to 1.17 points higher than a single pathologist's across retention levels; before referral the model's 6.96 points lay within the range of the three pathologists (4.24 to 8.29). In the score-substitution simulation, the combined score at 90% retention had slightly higher MAE than a pathologist scoring all slides, although the confidence interval included zero (6.69 versus 6.01 percentage points; difference of −0.68 for pathologist minus score substitution, 95% CI −1.93 to +0.57). At 70% retention, the confidence interval for the difference also included zero (6.37 versus 6.01; difference −0.36, 95% CI −0.93 to +0.18). Intervals use 2,000 WSI bootstrap samples with the retention decisions held fixed.

## S8. Largest Residuals and Referral Orderings

| Rank | WSI | Specimen | Three-reader mean TPS | D2Count-Net TPS | Abs. error | Likely failure mode |
|---|---|---|---|---|---|---|
| 1 | 400 | EBUS | 86.67 | 31.78 | 54.89 | high-TPS under-scoring; EBUS/small-specimen context |
| 2 | 327 | Non-EBUS | 54.00 | 2.90 | 51.10 | high-TPS under-scoring |
| 3 | 218 | Non-EBUS | 61.67 | 11.25 | 50.41 | high-TPS under-scoring |
| 4 | 328 | Non-EBUS | 47.33 | 8.18 | 39.15 | large TPS residual |
| 5 | 247 | Non-EBUS | 65.00 | 29.77 | 35.23 | high-TPS under-scoring; detection–density conflict |
| 6 | 138 | EBUS | 63.33 | 30.80 | 32.53 | high-TPS under-scoring; detection–density conflict; EBUS/small-specimen context |
| 7 | 120 | Non-EBUS | 38.33 | 6.00 | 32.34 | large TPS residual |
| 8 | 243 | EBUS | 38.33 | 9.45 | 28.88 | large TPS residual; EBUS/small-specimen context |
| 9 | 392 | Non-EBUS | 32.33 | 3.50 | 28.84 | large TPS residual |
| 10 | 109 | Non-EBUS | 30.00 | 5.56 | 24.44 | large TPS residual |

**Supplementary Table S5.** Ten largest differences between the model and the three-reader mean among the 114 WSIs without cell-level annotations, with specimen group and likely failure mode. Failure modes were assigned by fixed rules: high-TPS under-scoring when the three-reader mean was 50% or more and the model score was lower (otherwise, large TPS residual); detection–density conflict when the detection-branch TPS exceeded the density-branch TPS by more than 25 percentage points; and EBUS/small-specimen context for EBUS specimens. EBUS, endobronchial ultrasound.

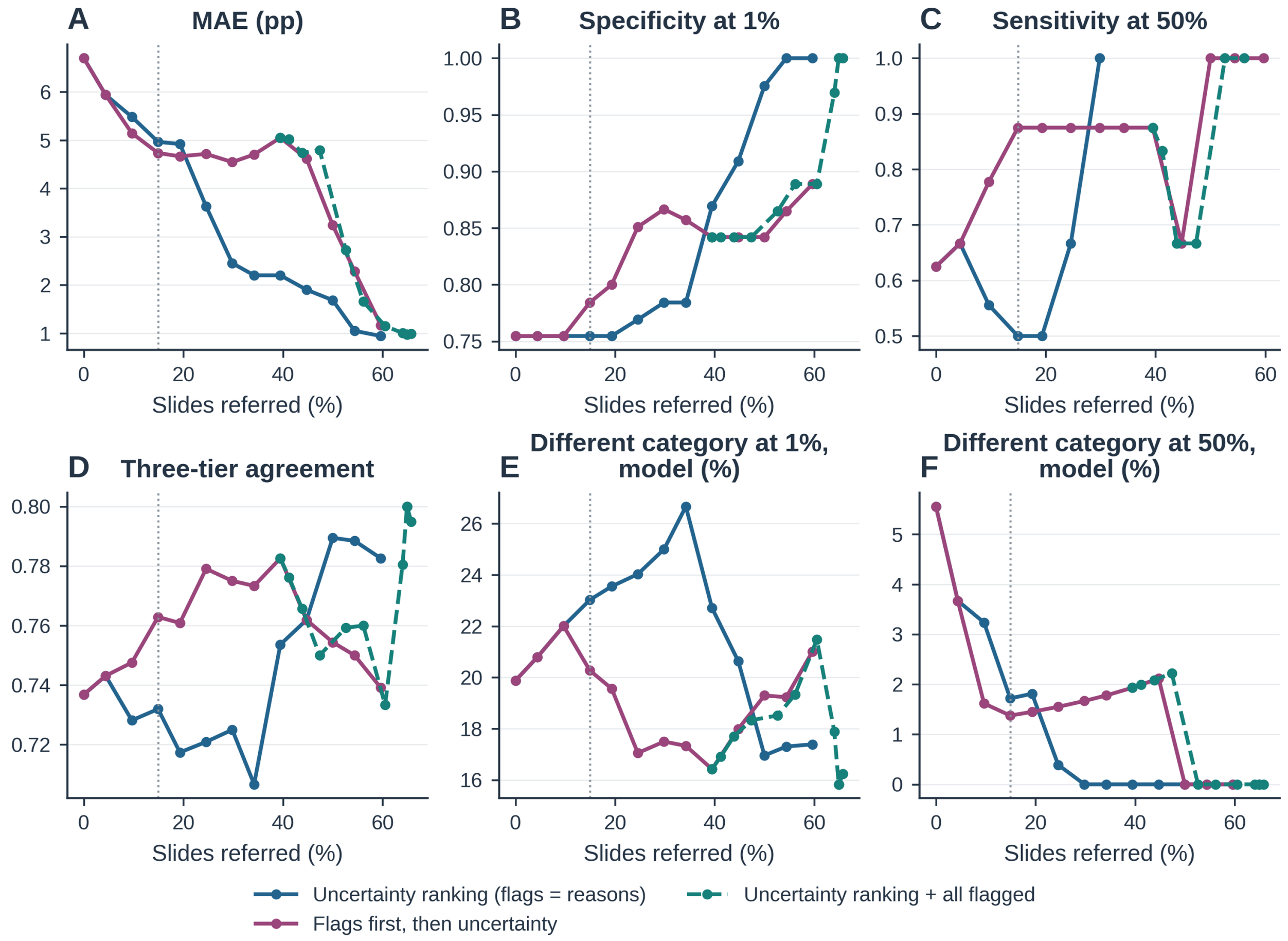

**Supplementary Figure S5.** TPS-Guard referral orderings on the 114 WSIs without cell-level annotations. Metrics among auto-reported slides are shown against the percentage of slides referred for three orderings: uncertainty ranking with flags as reasons (the main analysis), flags first and then uncertainty, and uncertainty ranking with all flagged slides also referred (dashed; it starts at the rate set by the number of flagged slides). The dotted line marks 15% referred. (A) MAE. (B) Specificity at the 1% cutoff. (C) Sensitivity at the 50% cutoff. (D) Three-tier agreement. (E, F) Auto-reported slides placed in a different category from the reference at the 1% and 50% cutoffs; each pathologist is held out in turn, the model is compared with the mean of the other two readers, and values are averaged over the three hold-outs. (A–D) use the three-reader mean.

## S9. Specimen-Stratified Agreement by Evaluation Frame

| WSIs | Specimen | n | Pathologist pairs | D2Count-Net vs three-reader mean | Difference | Single pathologist vs other two | D2Count-Net vs other two | Difference |
|---|---|---|---|---|---|---|---|---|
| All 154 WSIs | EBUS | 25 | 0.867/9.61 | 0.843/10.33 | −0.024/−0.72 | 0.903/9.23 | 0.837/10.56 | −0.066/−1.33 |
| All 154 WSIs | Non-EBUS | 129 | 0.880/5.45 | 0.874/5.82 | −0.005/−0.37 | 0.908/4.99 | 0.865/6.06 | −0.043/−1.07 |
| 114 WSIs without cell-level annotations | EBUS | 19 | 0.852/10.26 | 0.828/10.98 | −0.023/−0.72 | 0.892/9.88 | 0.822/11.23 | −0.070/−1.36 |
| 114 WSIs without cell-level annotations | Non-EBUS | 95 | 0.860/5.63 | 0.869/5.84 | +0.009/−0.21 | 0.893/5.23 | 0.859/6.10 | −0.034/−0.87 |

**Supplementary Table S6.** Endobronchial ultrasound (EBUS)-guided biopsy versus non-EBUS subgroup agreement across the two evaluation frames defined in Methods Section 1.3. Non-EBUS includes surgical resections, wedge biopsies, and the single lymph node specimen. The 114 WSIs exclude all WSIs with cell-level annotations. In these 114 WSIs, D2Count-Net MAE before referral was 6.70 percentage points, compared with pairwise pathologist MAE of 6.41. When the model and each held-out pathologist were both compared with the mean of the other two readers, the model's error exceeded a single pathologist's by 0.95 to 1.12 points across evaluation frames. Agreement entries are ICC/MAE, with MAE in TPS percentage points. ICC, intraclass correlation coefficient; MAE, mean absolute error. In the last three columns the model and each pathologist are compared with the mean of the other two readers. Each difference is model minus pathologist for ICC and pathologist minus model for MAE. Positive values favor the model.

## S10. Ensemble Uncertainty and Expert Disagreement

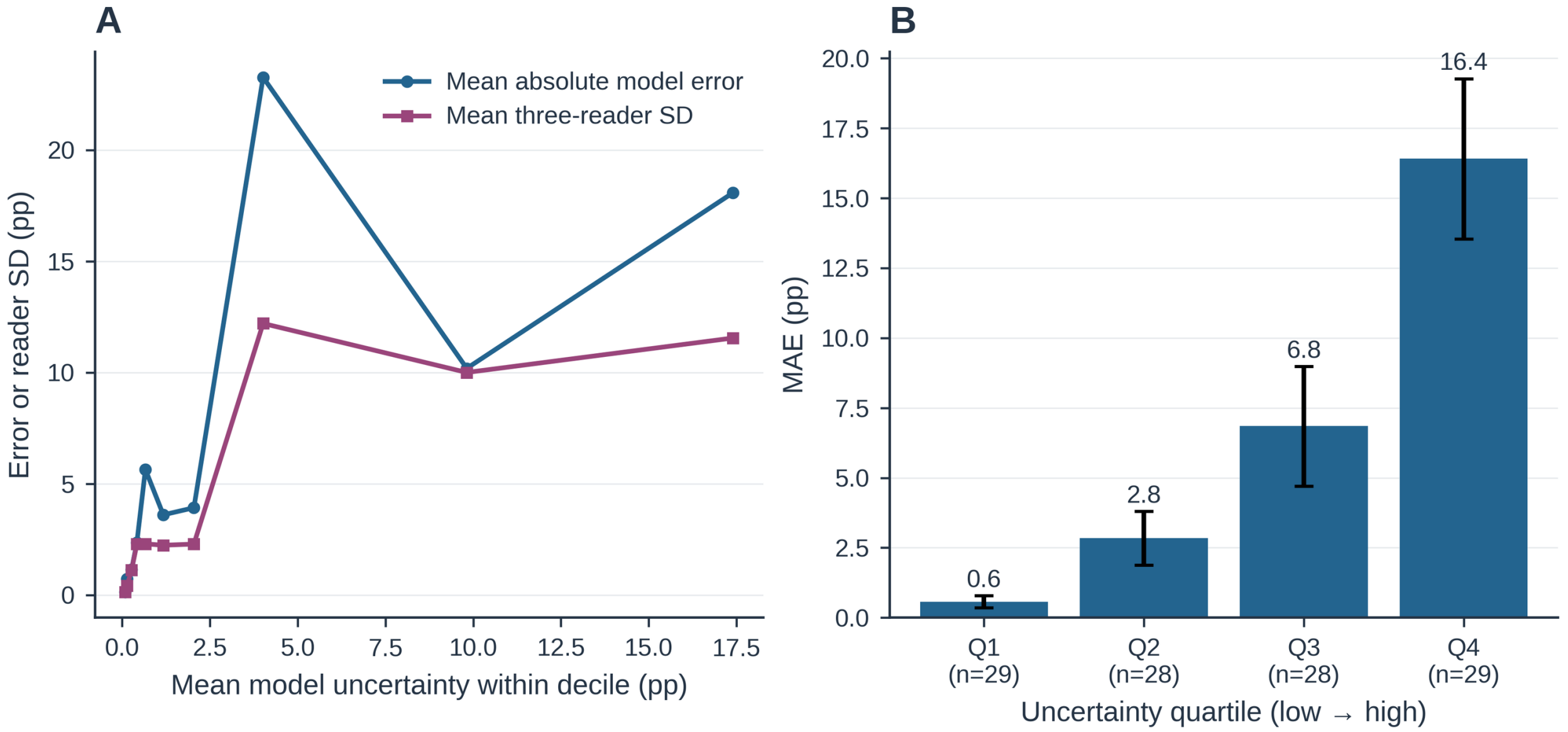

**Supplementary Figure S6.** Model error and reader disagreement across levels of model uncertainty on 114 WSIs. (A) Mean absolute TPS error and mean three-reader SD in ten approximately equal-sized groups ordered by model uncertainty (11–12 slides per group). Points are plotted at each group's mean uncertainty. (B) Mean absolute TPS error by uncertainty quartile, from Q1 (lowest) to Q4 (highest), with 28–29 slides per quartile. Error bars show the standard error of the mean. Model uncertainty is the SD of the TPS estimates from the 15 models; reader disagreement is the population SD of the three TPS scores. The individual-slide association is shown in main Figure 5C and 5D. SD, standard deviation; pp, percentage points.

## S11. Representative Cell-Counting Outputs for External ROIs

Supplementary Figure S7 compares cell count estimates for one TPS-negative external ROI under three conditions: application without retraining, external-only training, and fine-tuning. The development models underestimated the total cell count when applied without retraining. The fine-tuned detection branch estimated approximately 60 PD-L1-negative tumor cells and no false-positive cells, compared with 55 annotated negative cells and no annotated positive cells. Density and fusion overestimated cell counts in this example. Main Table 3 and Supplementary Figure S8 report performance across the external test cohort.

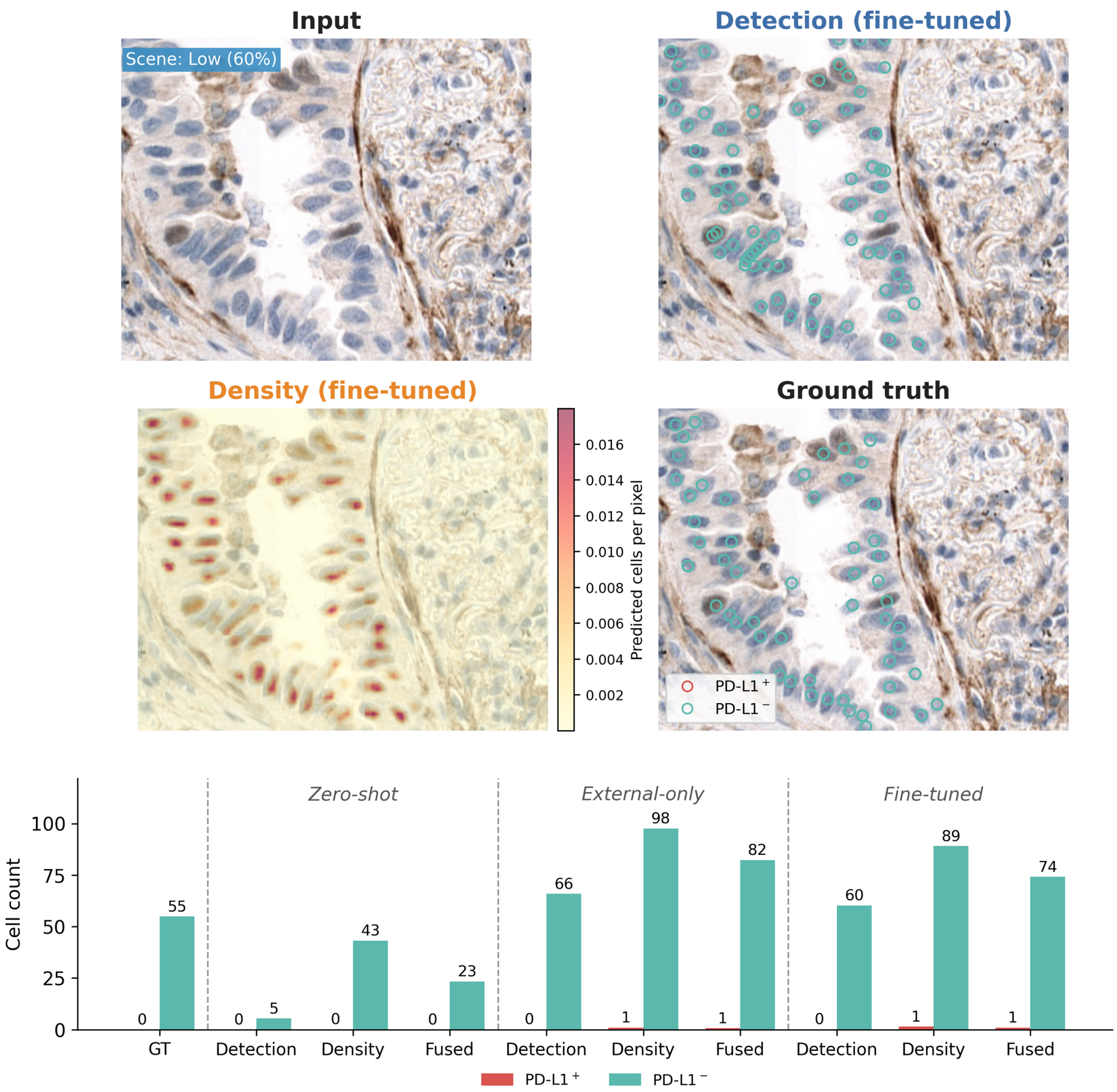

**Supplementary Figure S7.** Selected TPS-negative external ROI illustrating cell count estimates before and after domain adaptation. Top, input ROI and fine-tuned detection-branch overlay. Middle, fine-tuned density-branch heatmap and annotated tumor cells (red, PD-L1-positive; teal, PD-L1-negative). Bottom, positive and negative cell count estimates from application without retraining (labeled "Zero-shot"), external-only training from ImageNet initialization, and fine-tuning. The first and third conditions use the 15-model ensembles; external-only training uses its ensemble of five models. Bars show counts summed over the 224 × 224-pixel patches covering the ROI for each model, then averaged across the models. The number of detection circles can differ from the bar value because circles mark local maxima in averaged prediction maps and bars report averaged counts. Bar labels round the continuous estimates to integers. The input-panel label is the predicted cellularity scene and its model probability. Density is computed per tile, leaving visible tile boundaries.

## S12. External Validation Summary

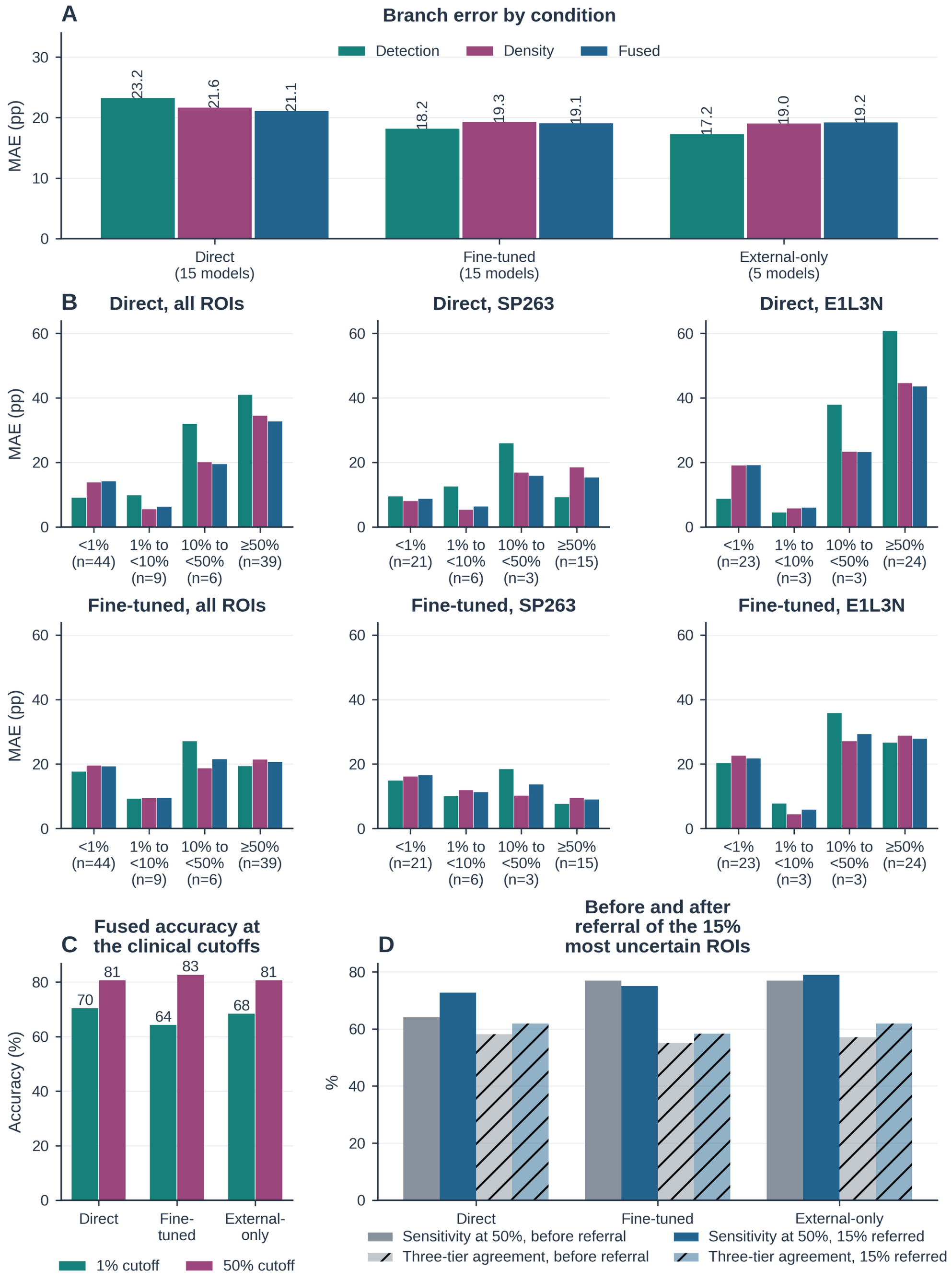

**Supplementary Figure S8.** External validation on the IGNITE cohort. (A) MAE by branch for the three conditions: the 15-model ensemble applied without retraining (direct), the fine-tuned 15-model ensemble and the external-only 5-model ensemble. (B) MAE by reference TPS range and branch for direct application (top row) and the fine-tuned ensemble (bottom row), in all ROIs and in the SP263 and E1L3N subsets; n, number of ROIs in each range. (C) Fused accuracy at the 1% and 50% cutoffs. (D) Sensitivity at the 50% cutoff and three-tier agreement before referral and after referral of the 15% most uncertain ROIs, for the three conditions.

### S13. External Clone-Subgroup Analysis

TPS-Guard outputs at 85% target retention are stratified by PD-L1 antibody clone for the 15-model ensemble applied without retraining. Results are reported separately for SP263 (45 ROIs, the clone shared with the development cohort) and E1L3N (53 ROIs, not represented in development training). Uncertainty rankings were recalculated within each clone, with a separate 85% retention target for each clone; for the flag-based curves in Supplementary Figure S9, count quantiles and flags were also recalculated within each clone.

*Panel A. 15-model ensemble applied without retraining.*

| Clone | Configuration | Actual retention | MAE (pp) | ICC | Specificity at 1% | Sensitivity at 50% | Three-tier discordance |
|---|---|---|---|---|---|---|---|
| SP263 (n = 45) | Uncertainty ranking (main analysis) | 0.867 | 9.01 | 0.900 | 0.667 | 0.923 | 0.282 |
| SP263 | No referral | 1.000 | 11.06 | 0.878 | 0.571 | 0.867 | 0.333 |
| E1L3N (n = 53) | Uncertainty ranking (main analysis) | 0.868 | 28.80 | 0.498 | 0.455 | 0.571 | 0.478 |
| E1L3N | No referral | 1.000 | 29.66 | 0.473 | 0.435 | 0.500 | 0.491 |

*Panel B. Auto-reported counts and cutoff classifications.*

| Clone / referral | Auto-reported n | Sensitivity at 50% TP/(TP + FN) | Specificity at 1% TN/(TN + FP) |
|---|---|---|---|
| SP263 Uncertainty ranking | 39 | 12/13 | 12/18 |
| SP263 No referral | 45 | 13/15 | 12/21 |
| E1L3N Uncertainty ranking | 46 | 12/21 | 10/22 |
| E1L3N No referral | 53 | 12/24 | 10/23 |

**Supplementary Table S7.** Per-clone TPS-Guard performance at 85% target retention. (A) 15-model ensemble applied without retraining. (B) Auto-reported counts and cutoff classifications. Uncertainty ranking is the main analysis. Sensitivity uses reference-positive ROIs at 50%; specificity uses reference-negative ROIs at 1%. TP, true positive; FN, false negative; TN, true negative; FP, false positive. SP263 (45 ROIs) is the development clone; E1L3N (53 ROIs) was absent from development training.

At 85% within-clone target retention, ranking by model uncertainty alone gave MAE of 28.80 percentage points on E1L3N versus 9.01 on SP263. On E1L3N, referral increased the sensitivity of auto-reported ROIs at the 50% cutoff from 0.50 to 0.57 and reduced three-tier discordance from 0.49 to 0.48. Substantial TPS error remained on E1L3N after referral.

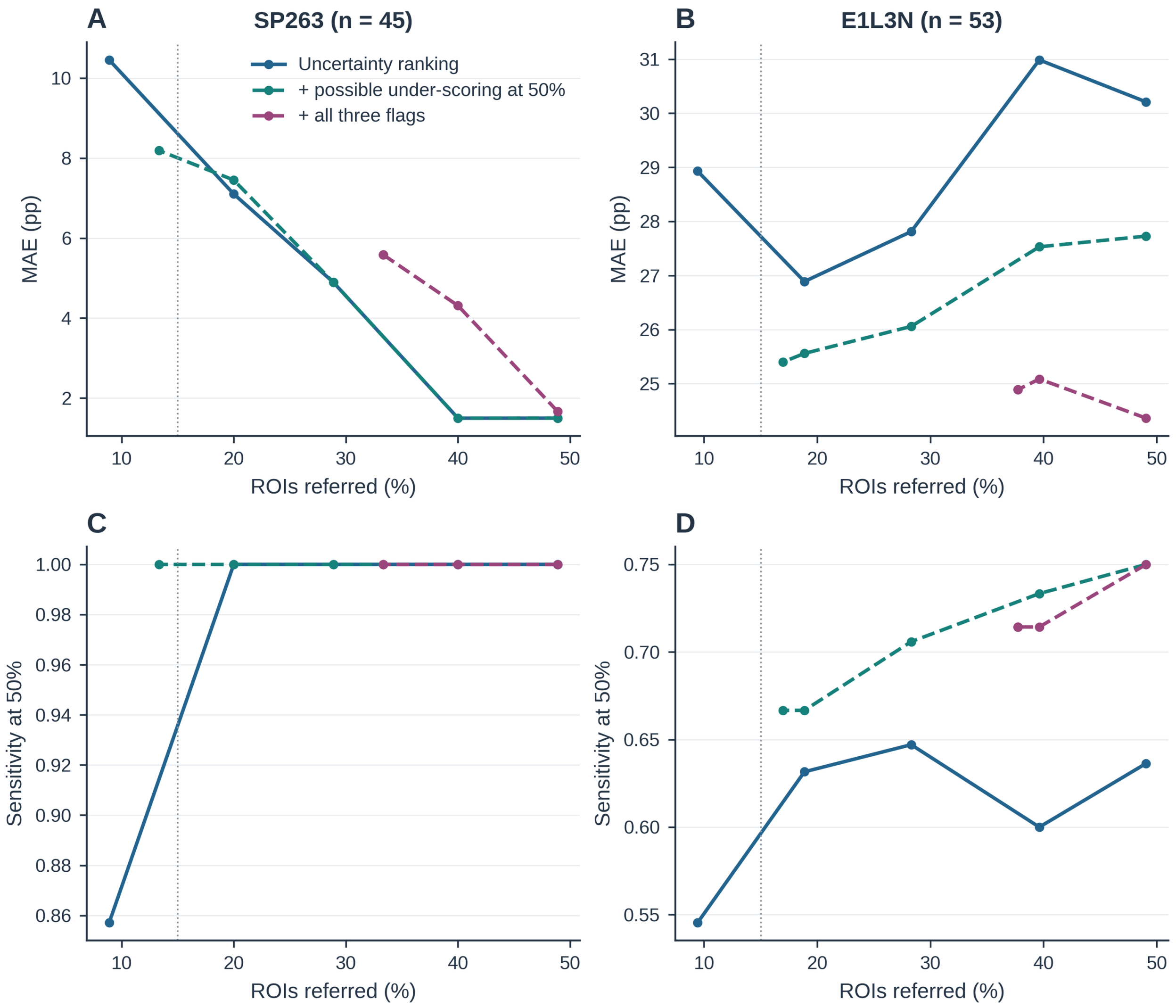

**Supplementary Figure S9.** External clone-subgroup referral curves (15-model ensemble applied without retraining). Solid lines, uncertainty ranking; dashed lines, ROIs flagged for possible under-scoring at 50% (or with any flag) are referred first and the remaining referral quota is filled by uncertainty ranking; the dotted line marks 15% referred. (A, B) MAE of auto-reported ROIs against the percentage of ROIs referred on the SP263 subset ($n = 45$ ROIs) and the E1L3N subset ($n = 53$ ROIs). (C, D) Sensitivity at the 50% cutoff on the same two subsets.

## S14. Context with Prior Automated PD-L1 TPS Systems

| Reference | Cohort | Ground truth | Method | Reported metric |
|---|---|---|---|---|
| Kapil et al.[9] | 40 test NSCLC needle biopsies (SP263) | Median of three visual TPS assessments | AC-GAN semi-supervised region segmentation; TPS from positive-to-total tumor area | Pearson r = 0.95; MAE = 8.00 percentage points |
| Pan et al.[14] | 192 SP263-stained and 117 22C3-stained samples from a cohort of 230 samples | Pathologist visual TPS | Multistage ensemble (regional segmentation + cellular detection) | ICC = 0.873 for SP263; ICC = 0.737 for 22C3 |
| Van Eekelen et al.[12] | 100 NSCLC WSIs; multicenter, multi-assay reader study | Six readers; paired reader–AI comparisons | YOLOv5 cell detection | Reader–AI ICC = 0.796 (continuous TPS); Cohen's κ = 0.494 (TPS categories) |
| Present work (UHCW) | 154 NSCLC WSIs, single center, single clone (SP263) | Mean TPS from three independent pathologists | D2Count-Net with TPS-Guard triage | ICC 0.869 vs the three-reader mean on the full cohort of 154 WSIs and 0.862 on the 114 WSIs without cell-level annotations; in EBUS slides without cell annotations (n = 19), leave-one-reader-out ICC/MAE 0.822 / 11.23 versus a single pathologist 0.892 / 9.88 |

**Supplementary Table S8.** Cohort, ground truth, method, and reported metric for selected automated PD-L1 TPS scoring systems and the present work. NSCLC, non-small cell lung cancer; AC-GAN, auxiliary classifier generative adversarial network; AI, artificial intelligence.

Supplementary Table S8 summarizes selected scoring systems. Differences in cohort, reference standard, and metric limit direct numerical comparisons. Makhlouf et al.[16] also described cutoff-related manual review in PD-L1 assessment. The present study adds an evaluation of model-derived risk signals, their association with reader disagreement, and the reasons assigned to referred cases.

## S15. Positive-Class Reweighting

To test whether the small number of annotated PD-L1-positive tumor cells could be offset by reweighting, the 15 models were retrained with the PD-L1-positive channel of the detection-map loss weighted by 2.4 (the square root of the ratio of annotated PD-L1-negative to PD-L1-positive tumor cells), with everything else unchanged and the rule fixed before training. On the 114 WSIs without cell-level annotations, MAE was unchanged (6.70 versus 6.70 percentage points), bias for slides with a three-reader mean of 20% to <50% and 50% to <80% moved toward zero by 1.0 and 1.7 percentage points, and three-tier agreement fell from 0.737 to 0.711 (Supplementary Table S9). Results on all 154 WSIs and at the patch level showed the same pattern. The reweighted models were not adopted.

| Metric | Reference models | Reweighted models |
|---|---|---|
| MAE (pp) | 6.70 | 6.70 |
| ICC | 0.862 | 0.866 |
| Bias, all slides (pp) | −4.96 | −4.58 |
| Bias, reader mean 20% to <50% (pp; n = 13) | −18.99 | −17.99 |
| Bias, reader mean 50% to <80% (pp; n = 9) | −22.21 | −20.51 |
| Three-tier agreement | 0.737 | 0.711 |
| Specificity at 1% | 40/53 | 38/53 |
| Sensitivity at 50% | 10/16 | 10/16 |
| Leave-one-reader-out MAE, model (pathologist 6.01) | 6.96 | 6.91 |

**Supplementary Table S9.** Reference and positive-class reweighted models on the 114 WSIs without cell-level annotations, before referral. Bias is model minus three-reader mean. MAE, mean absolute error; ICC, intraclass correlation coefficient; pp, percentage points.